\documentclass[journal]{IEEEtran}
\ifCLASSINFOpdf
\else
\fi
\usepackage{amsmath}
\usepackage{amssymb}
\usepackage{adjustbox}
\usepackage{xcolor}
\usepackage{color}
\usepackage{flushend}
\usepackage{hyperref}

\definecolor{green(html/cssgreen)}{rgb}{0.0, 0.5, 0.0}

\newcommand{\ov}[1]{\textbf{\textcolor{orange}{#1}}}
\newcommand{\bv}[1]{\textbf{\textcolor{blue}{#1}}}
\newcommand{\gv}[1]{\textbf{\textcolor{green(html/cssgreen)}{#1}}}
\usepackage{array}
\newcolumntype{P}[1]{>{\centering\arraybackslash}p{#1}}

\usepackage{graphicx}
\ifCLASSOPTIONcompsoc
  \usepackage[caption=false,font=normalsize,labelfont=sf,textfont=sf]{subfig}
\else
  \usepackage[caption=false,font=footnotesize]{subfig}
\fi
\begin{document}
%
\title{Leveraging Ensembles and Self-Supervised Learning for Fully-Unsupervised Person Re-Identification and Text Authorship Attribution}
%
%
%

\author{Gabriel~Bertocco,
        Antonio~Theophilo, 
        Fernanda~Andal\'{o},~\IEEEmembership{Member,~IEEE,}
        and~Anderson~Rocha,~\IEEEmembership{Senior~Member,~IEEE}
\thanks{Gabriel Bertocco is a Ph.D. student at the Artificial Intelligence Lab. (\textbf{Recod.ai}), Institute of Computing, University of Campinas, Brazil}
\thanks{Antonio~Theophilo is a Ph.D. student at the Artificial Intelligence Lab. (\textbf{Recod.ai}), Institute of Computing, University of Campinas, Brazil, and a researcher at the Center for Information Technology Renato Archer (CTI) Campinas, Brazil.}

\thanks{Fernanda~Andal\'{o} is a researcher associated with the Artificial Intelligence Lab. (\textbf{Recod.ai}), Institute of Computing, University of Campinas, Brazil.}
\thanks{Anderson Rocha is an Associate Professor and Chair of the Artificial Intelligence Lab. (\textbf{Recod.ai}) at the Institute of Computing, University of Campinas, Brazil. 
}
\thanks{This paper has supplementary downloadable material available at http://ieeexplore.ieee.org., provided by the author. The material includes a manuscript with additional experimental results and image visualizations. Contact gabriel.bertocco@ic.unicamp.br for further questions about this work.}}

\maketitle

\begin{abstract}
Learning from fully-unlabeled data is challenging in Multimedia Forensics problems, such as Person Re-Identification and Text Authorship Attribution. Recent self-supervised learning methods have shown to be effective when dealing with fully-unlabeled data in cases where the underlying classes have significant semantic differences, as intra-class distances are substantially lower than inter-class distances. However, this is not the case for forensic applications in which classes have similar semantics and the training and test sets have disjoint identities. General self-supervised learning methods might fail to learn discriminative features in this scenario, thus requiring more robust strategies. We propose a strategy to tackle Person Re-Identification and Text Authorship Attribution by enabling learning from unlabeled data even when samples from different classes are not prominently diverse. We propose a novel ensemble-based clustering strategy whereby clusters derived from different configurations are combined to generate a better grouping for the data samples in a fully-unsupervised way. This strategy allows clusters with different densities and higher variability to emerge, reducing intra-class discrepancies without requiring the burden of finding an optimal configuration per dataset. We also consider different Convolutional Neural Networks for feature extraction and subsequent distance computations between samples. We refine these distances by incorporating context and grouping them to capture complementary information. Our method is robust across both tasks, with different data modalities, and outperforms state-of-the-art methods with a fully-unsupervised solution without any labeling or human intervention.

\end{abstract}


%
\IEEEpeerreviewmaketitle

\section{Introduction}
\label{sec:introduction}
%
%
%
%

\IEEEPARstart{S}{elf}-supervised learning has been gaining increasing attention due to its capacity to learn patterns from unlabeled data. Techniques for self-supervised learning usually rely on contrastive strategies~\cite{caron2020unsupervised, he2020momentum, chen2020simple, caron2021emerging}, where augmented views of the same image are arranged to be closer in an embedding space but further apart from other images. 

It is common practice to consider ImageNet~\cite{deng2009imagenet} as the dataset of choice for experimentation, as its high number of classes usually translates to general features, providing a good weight initialization for transfer learning. 
Although ImageNet contains some closer classes (e.g., different dog breeds), most of them have very different semantics (e.g., cars, airplanes, fruits, and flowers) and are easier to distinguish. This enables the learning process to rely more on coarse rather than fine-grained details, and the mere use of well-known augmentation strategies (such as cropping, blurring, erasing, flipping) provides sufficient variation for optimization~\cite{caron2020unsupervised, he2020momentum, chen2020simple}. Nonetheless, coarse features are insufficient for several critical tasks, requiring fine-grained focus and other strategies to enable effective self-supervised learning. Examples of critical tasks are Person Re-Identification (ReID), and Text Authorship Attribution. 

Person ReID aims to retrieve the same person seen in one camera from all the other cameras present in a system. In this scenario, all classes are semantically identical as they represent people in the same environment. Therefore, the model must overcome point-of-view changes, occlusions, different lighting, resolutions, and backgrounds. Moreover, people in a specific place tend to dress similarly. DukeMTMC-ReID~\cite{ristani2016performance}, a ReID dataset composed of videos from eight cameras at a university campus during winter, presents most of the identities wearing pants, dark coats, backpacks, boots, and other attributes that people usually wear in cold weather or universities. Therefore, most identities (classes) tend to be similar, which means a high inter-class similarity. In contrast, a person might be recorded from different angles by different cameras, which results in a higher intra-class disparity.

Moreover, Person ReID is intrinsically an open-world problem: identities in the test set (query and gallery sets) are disjoint from those in the training set, while ImageNet training and test sets have the same classes. Aside from this, ReID datasets can be as large as ImageNet: Market1501~\cite{zheng2015scalable} has 751 identities in the training set, DukeMTMC-ReID~\cite{ristani2016performance} has 702, and MSMT17~\cite{wei2018person} even exceeds ImageNet with 1041 classes in the training set. This clearly shows that Person ReID is a more demanding task than ImageNet classification, and general state-of-the-art self-supervised learning methods are not suitable for the task. Previous works~\cite{ge2020self, zhang2021refining} verified that these methods result in much less accurate models when compared to unsupervised learning methods tailored specifically to ReID, even with ImageNet weight initialization. 

Text Authorship Attribution (TAA) faces similar challenges. TAA aims to recover, from a gallery, texts from the same author of a query text. Multiple authors writing a social media post about the same topic (e.g., politics, sports, economics, etc.) might use similar vocabulary, which can make the texts similar, resulting in high inter-class similarity. Conversely, the same author can write about different topics, which leads to high intra-class disparity. Moreover, we address a more challenging scenario than the usual evaluation scenario for TAA, since we adopt the open-set scenario where training and test sets are disjoint in terms of identities.

In this context, we propose a novel self-supervised learning approach to handling the fully-unsupervised Person ReID and Text Authorship Attribution tasks, which require a robust distance measure and a fine-grained analysis. We start by considering a common approach: \textbf{clustering steps} to propose pseudo-labels to unlabeled samples and \textbf{optimization steps} to update the model supervised by those pseudo-labels~\cite{ge2020self, chen2020enhancing, 9521886}. However, prior methods that consider this approach often overlook two aspects: the quality of the features and the choice of hyper-parameters for the clustering algorithm. If the features are not too descriptive, samples from different classes might end up closer in the feature space, leading them to be clustered together, increasing the number of false positives and ultimately hindering model updates. Even when the features are adequate, a bad choice of hyper-parameters for the clustering process might yield suboptimal groups.

To address these problems, we take inspiration from re-ranking techniques~\cite{zhong2017re} to filter out false-positive samples. The proposed method starts by calculating pairwise distances for unlabeled samples, considering features extracted by $M$ Deep Neural Networks (we refer to them as \textit{backbones}). Those distances are normalized by considering the mutual neighbors in each of the $M$ feature spaces. As a second distance refinement, we average the $M$ distances between two samples, as each backbone can provide a complementary description. Recent works consider ensemble techniques for Unsupervised Domain Adaptation~\cite{ge2020mutual, zhai2020multiple}. These studies, however, apply mutual learning by leveraging complex loss functions with one backbone supervising the other, which brings complexity to the training process. Our method, in turn, ensembles models by only taking the average distances, allowing the amalgamation of complementary information from each manifold, but with a much more straightforward setup.

The second aspect is the choice of hyper-parameters for the clustering process. We take DBSCAN as our clustering algorithm and, instead of fixing a value for the $\varepsilon$ parameter, we scan different clustering densities --- the lower the value, the denser the cluster. If a sample is identified as an outlier in any of these levels, it is marked as an outlier in the other levels. By tracking the state of each sample (inliers and outliers) through clustering runs, we can produce clusters that connect different dense regions while disregarding noisy samples. 

Our method is designed for two critical applications that operate with different types of data: Person re-identification (images) and Authorship Attribution for social media (texts). 
To the best of our knowledge, we are the first to apply the same self-supervised learning method to different modalities with minor adjustments. 

The main contributions of our work are:
\begin{itemize}
    \item An effective distance averaging strategy to combine distances between feature vectors generated by independent backbones, taking advantage of complementary information. This simplifies the training process, as previously proposed complex techniques are unnecessary.
    \item An ensemble-based clustering strategy in which we scan a set of hyper-parameter values and combine intermediate clustering results in a unique, more robust result. By doing this, we can connect dense regions without the effect of noisy samples.
    \item A novel self-supervised learning formulation that can be applied to different problems such as fully-unsupervised Person ReID and Text Authorship Attribution with minor adjustments.
\end{itemize}

\section{Related Work}

In this section, we describe related methods for self-supervised learning, with a more detailed exploration of person re-identification and text analysis.

\subsection{Self-Supervised Learning}
Self-Supervised learning is usually done by generating two or more views of the same sample through augmentation techniques. A contrastive loss is minimized to pull together different views of the same image while pushing original images of different classes apart.

MoCo~\cite{he2020momentum} generates two random augmented versions for each image of a batch. One is fed to a key encoder, and the other to a query encoder for feature extraction. The features from the key encoder are added to a dictionary that stores features from previous batches. The method minimizes a contrastive loss to pull together both augmented versions while keeping other features in the dictionary apart. 
SimCLR~\cite{chen2020simple} also adopts two augmented versions of each image, but without a dictionary of features. Instead, it minimizes a contrastive loss by considering both augmented images as a positive pair and the other images (and their augmented versions) as negative. 

SWaV~\cite{caron2020unsupervised} applies a multi-cropping strategy by considering two standard-resolution crops and several low-resolution crops for optimization. Like SWaV, Dino~\cite{caron2021emerging} adopts different levels of cropping to optimize a teacher-student loss function. They feed the global crops to the teacher network to get a final probability distribution used to supervise the student network, which is fed with local crops. 

These methods obtain competitive performance when compared to their supervised learning equivalents. However, as they are generally tested on ImageNet, they tend to fail on problems with high intra-class dissimilarity and inter-class similarity, such as person ReID and Authorship Attribution of short messages.

\subsection{Unsupervised Person Re-Identification}

To tackle unsupervised person re-identification, some methods rely on pre-training a model on a source ReID dataset to acquire prior knowledge of the problem. This model is then adapted to the unlabeled target domain.

ECN-GPP~\cite{zhong2020learning} is a pipeline that handles example-, camera-, and neighborhood-invariance to enforce images to be closer to their closest samples, by minimizing all three aspects in a loss function. 
DG-Net++~\cite{zou2020joint} proposes a pipeline to disentangle ID-related features by generating images on target and source domains with semantic features across both domains. They also consider a clustering algorithm to propose pseudo-labels to unlabeled data to regularize training. 

MMT~\cite{ge2020mutual} and MEB-Net~\cite{zhai2020multiple} are ensemble-based methods that leverage two and three backbones, respectively, one supervising the other in a teacher-student regime. They propose soft and hard pseudo-labels to the unlabeled samples to perform optimization. 
ABMT~\cite{chen2020enhancing} leverages a teacher-student model, where the global average pooling branch supervises the global max pooling branch, and vice-versa.
In a previous work~\cite{9521886} of ours, we considered ensembles only during evaluation. We generated cross-camera triplets using camera information of samples in the generated clusters. We also proposed a self-ensembling strategy in which the training of each backbone is summarized by weight averaging the checkpoints.

Instead of considering pre-training in the ReID domain, another set of methods relies on other pieces of information, such as camera labels. 
IICS~\cite{xuan2021intra} leverages intra-camera training by dividing samples into sets according to their camera labels and performing clustering on each one. A backbone is trained in a multi-task manner (one task per camera), and clustering is run for the whole dataset, grouping samples of the same identity seen from different cameras. 
CAP~\cite{wang2020camera} performs global clustering by assigning pseudo-labels for each sample of the dataset. They obtain camera proxies on each cluster for intra- and inter-camera training. 
ICE~\cite{Chen_2021_ICCV} has two versions: camera-aware and camera-agnostic. The first one considers the camera proxy features similar to CAP. The second considers only the cluster proxy, obtained by averaging features regardless of the camera label. They use a proxy-based loss along with a hard- and a soft-instance loss. 

Instead of camera labels, some works rely on tracklets. 
CycAs~\cite{wang2020cycas} aims to identify the same person in frames of a video for intra-sampling. They also find the same person in other videos by considering the overlapping field-of-view between two videos for inter-sampling. With both intra- and inter-sampling, they optimize the backbone to match the same person from different points of view.
UGA~\cite{wu2019unsupervised} averages the features of the same person in the same tracklet and performs cross-camera feature association creating a Cross-View-Graph (CVG) to encourage the matching of tracklets of the same person from different points of view. 

Other works assume that only person bounding boxes are available. These are considered fully unsupervised.
ABMT~\cite{chen2020enhancing} relies on source pre-training, but the authors also present results when no prior knowledge is considered. 
SpCL~\cite{ge2020self} proposes a self-paced strategy that introduces metrics to measure cluster reliability: cluster independence and cluster compactness. If both are higher than predefined thresholds, the cluster is kept within the feature space. They also minimize the loss function considering cluster centroids and samples stored in feature memory. 
RLCC~\cite{zhang2021refining} refines clusters by a consensus among iterations. Pseudo-labels on a certain iteration are created by considering the ones generated on a previous iteration, keeping the training stable. CACL~\cite{li2022cluster} proposes a strategy to suppress the dominant colors on images, providing a more robust feature description, and a novel pseudo-label refinement method. ISE~\cite{zhang2022implicit} synthesizes novel feature examples from real ones to refine the sample distribution, aiming to generate clusters with a higher true positive rate, as well as avoiding subdivision of the samples from the same identity in different clusters. PPLR~\cite{cho2022part} employs a part-based model that creates feature spaces from different parts of the feature map. In each one, they calculate the nearest neighbors of each sample and propose a cross-agreement metric to refine the proposed pseudo-labels.

Compared to such methods, our method also does not rely on any extra information, requiring only the bounding boxes of the people in the dataset; therefore, it best fits this last category of methods. Nonetheless, prior art often relies on clustering methods with manually chosen optimal hyper-parameter values, which might be impractical when working with unlabeled datasets. Our method differs from the rest by proposing a clustering criterion, which alleviates the burden of choosing optimal hyper-parameters. 

We summarize the pros and cons from the main Unsupervised Person Re-Identification works in Table~\ref{tab:comparison_prior_art}. The main advantage of our method is that it operates in a fully-unsupervised scenario without relying on dataset-specific clustering hyper-parameter tuning. It leverages ensemble-based feature spaces and clustering, and it is the only one that has been designed for multiple modalities (images and text).

\subsection{Unsupervised Text Analysis}

Text Analysis is another application that can be explored with unsupervised learning methods.

The Natural Language Processing (NLP) community witnessed a significant development with the introduction of models based on Attention and the Transformer architecture~\cite{vaswani2017attention}.

BERT~\cite{devlin2018bert} is one of the most successful models, applying an encoder-only architecture to solve many NLP tasks. The authors propose a self-supervised pre-training regime using masked language modeling and next-sequence prediction tasks, followed by a fine-tuning step using supervised data. Several works followed BERT, proposing variations using more targeted data. One example is BERTweet~\cite{nguyen2020bertweet}, in which the authors propose an extension to deal with tweets (short messages from Twitter).

T5~\cite{raffel2019exploring} takes a step forward and proposes a single architecture to solve any NLP problem that can be modeled as a text-to-text task. They apply the architecture presented by Vaswani et al.~\cite{vaswani2017attention} with small changes in the normalization, dropout, and embedding layers. 



Text Authorship Attribution (TAA) is the task of finding an author of a text solely by analyzing the textual information. We can reframe the task similar to ReID, in which we aim to recover, from a gallery, texts from the same author of a query text. Despite most of the research in TAA being in a closet-set scenario, in this work, we follow the more realistic and challenging scenario of open-set where the model may have never seen texts from the actual author, like in ReID. Despite current methods achieving good results for lengthy texts
, authorship attribution is still challenging for short texts~\cite{theophilo2019authorship}.

Nowadays, Authorship Attribution over social media data is an extremely compelling and relevant scenario involving textual information. Recently, Kirkpatrick~\cite{kirkpatrick2022who} presented an interesting episode that supports this statement and the problem's relevance in this paper's context. The author shows how two teams of researchers applied AI techniques over social media messages to find the authors of QAnon messages --- an anonymous creator of far-right political conspiracy theories. Despite the success of applying the Authorship Attribution techniques, this episode also showed how difficult and challenging the task is. The researchers dealt with a small set of 13 suspects, a considerable supervised dataset of 100,000 words from QAnon, and at least 12,000 words from each suspect. Furthermore, parallel investigations outside the textual universe reinforced the results (e.g., messages' timestamps from the suspects claiming they had discovered the QAnon existence).



In this work, we target the Authorship Authorship task in a fully-unsupervised way, considering a dataset of tweets (short text messages from the Twitter social media platform). We consider a challenging and less explored setup, in which the authors from the training set are unlabeled and disjoint from those in the test set. As our baseline, we consider AdHominem~\cite{boenninghoff2019explainable}, an attention- and LSTM-based model for Authorship Attribution originally trained in a supervised manner using social media posts. The pros and cons of AdHominem are highlighted in Table~\ref{tab:comparison_prior_art}.
\vspace{0.5cm}
\begin{table*}[htbp]
\tiny
\caption{Comparison of the prior art and our method in terms of pros and cons.}
\label{tab:comparison_prior_art}
\centering
\resizebox{1.0\textwidth}{!}{\begin{tabular}{|P{2.0cm}|P{7.0cm}|P{7.0cm}|}
\hline
Method & Pros & Cons \\ \hline

SSL~\cite{lin2020unsupervised} & Does not require clustering & Requires camera labels and relies solely on the $k$-NN of each image for positive samples mining  \\ \hline
CCSE~\cite{lin2020unsupervisedccse} & Performs clustering regularization to balance the number of samples in each clustering & Requires camera labels and artificial GAN-generated images that might include biases in model learning\\ \hline
MPRD~\cite{ji2021meta} & Does not require clustering and explores the neighborhood of image features through Graph Convolutional Networks & Requires camera labels for augmentation, and the optimization of two dependents networks (no decoupling)\\ \hline 
DSCE-MC~\cite{yang2021joint} & Leverages a symmetric and dynamic cross-entropy loss and a camera-based meta training & Requires the camera labels and nearest neighbors-based outlier reassignment which might introduce noise \\ \hline
JVTC~\cite{li2020joint} & Employs local and global view on loss function to regularize model learning & Requires camera labels, artificial GAN-generated images that might include biases in the learning, and frame annotation for temporal consistency calculation \\ \hline
JGCL~\cite{chen2021joint} & Generates images of the same person in different poses, which regularizes the model training & Requires camera labels and the training of a GAN together with the main backbone, which brings complexity to the training process\\ \hline
IICS~\cite{xuan2021intra} & Employs the AIBN to regularize model learning to achieve cross-view invariant features & Requires multi-task training per camera where the complexity grows depending on the number of cameras and pseudo-identities found per camera \\ \hline
CAP~\cite{wang2020camera} & Regularizes the model with intra- and inter-camera losses & Requires camera labels and multiple proxies per cluster to be constantly updated in an epoch\\ \hline
UST~\cite{9521886} & Leverages a cross-camera triplet creation and self-ensembling technique for checkpoints summary & Requires camera labels and an offline generation of all the triplets before training\\ \hline
ICE~\cite{Chen_2021_ICCV} & Employs a soft consistency loss to be robust to augmentation and has a camera-agnostic version & Leverages camera labels in its best version, presents a loss function with many hyper-parameters, and it shows results with a specific clustering $\epsilon$ for each dataset \\ \hline
PPLR~\cite{cho2022part} & Leverages a part-based guided label and loss refinement and presents a camera-agnostic version & Requires camera labels in the best version, and the part-based agreement and loss function are complex\\ \hline
Star-Dac~\cite{prasad2022spatio} & Provides an analysis about the time each identity is recorded by each camera and how they transit from one to another & Requires camera and timestamp annotation for each frame and a high-complex spatiotemporal-based clustering \\ \hline
TAUDL~\cite{li2018unsupervised} & Does not require clustering and performs cross-tracklet association in batch level & Requires camera labels and tracklet estimation, not considering any global view of the feature space\\ \hline
TSSL~\cite{wu2020tracklet} & Does not require camera labels and leverages a distribution-aware cluster distance for clustering & Assumes tracklet annotation but lacks the clustering for a global view \\ \hline
UTAL~\cite{li2019unsupervised} & Does not require clustering and leverages tracklet-based soft labels for learning & Requires camera and tracklet annotations, and the training complexity grows linearly with the number of cameras \\ \hline
CycAs~\cite{wang2020cycas} & Does not require clustering and leverages a self-adaptive temperature parameter & Requires camera and timestamp annotation, and information if two or more camera field of view overlap  \\ \hline
UGA~\cite{wu2019unsupervised} & Does not require clustering and leverages a Cross-View Graph for inter-camera tracklet association & Requires camera and tracking annotation, and a multi-task training that grows linearly with the number of cameras \\ \hline
BUC~\cite{lin2019bottom} & Does not require camera labels and leverages a diversity regularization term on clustering & Evaluates and gets the best checkpoint in the test set, which is unrealistic \\\hline
GPUFL~\cite{sun2021unsupervised} & Does not require clustering nor camera labels & Requires to keep $m$+1 memory banks, where $m$ is the number of parts of the feature, and only local-neighborhood mining is performed\\ \hline
MV-ReID~\cite{yin2021multi} & Does not require camera labels and leverages multi-patch optimization & Requires calculating the feature distance for each extracted patch, and the best checkpoint over a validation set is selected, which is unrealistic\\ \hline
MMCL~\cite{wang2020unsupervised} & Does not require camera labels nor clustering & Does not consider a global view of the feature space for positive and negative mining \\ \hline
HCT~\cite{zeng2020hierarchical} & Does not require camera labels & The best checkpoint over a validation set is selected, which is unrealistic, and the hyper-parameters might change for different datasets \\ \hline
ABMT~\cite{chen2020enhancing} & Exploits both global and max pooling for feature learning & The loss function is complex, with five terms \\ \hline
SpCL~\cite{ge2020self} & Does not require camera labels and reliable samples are selected through independence and compactness measures & Introduces more clustering hyper-parameters which might be challenging to tune for different datasets \\ \hline 
RLCC~\cite{zhang2021refining} & Does not rely on camera labels for learning and leverages pseudo-label consensus across iterations & Same as SpCL, and adds one more parameter for pseudo-label generation; however, it is sensitive to it \\ \hline
CACL~\cite{li2022cluster} & Does not rely on camera labels and performs ensemble learning & The best checkpoint is selected, which is unrealistic, and the clustering requires different optimal hyper-parameter values for each dataset \\ \hline
ISE~\cite{zhang2022implicit} & Does not rely on camera labels and uses sample extension & Leverages dataset-specific clustering hyper-parameters \\ \hline
AdHominem~\cite{boenninghoff2019explainable} & Predicts if the same author wrote two short-message posts in social media & Relies on fully \textbf{supervised} training \\
\hline
\textbf{Ours} & \textbf{Does not rely on camera labels, nor tracklets, and no hyper-parameters need to be tuned for clustering. It exploits ensembles and is the only one employed for both image and text-based tasks} & \textbf{Requires the training of more than one model, but it can be performed in parallel.} \\ \hline

\hline
\end{tabular}}
\end{table*}

\section{Proposed Method}
\label{sec:proposed_method}

\begin{figure*}[ht]
\centering
\includegraphics[width=4.5in]{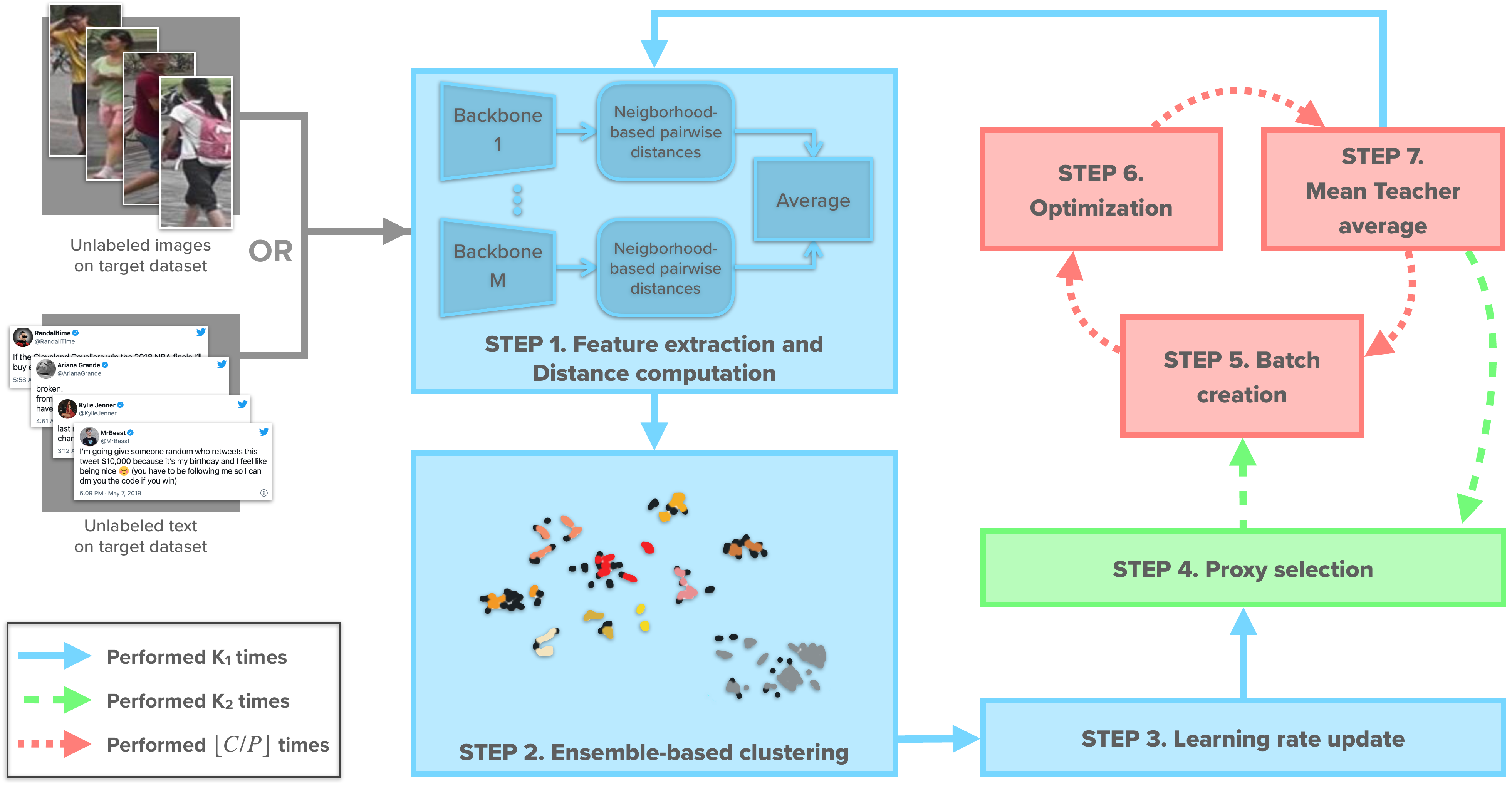}
\caption{Overview of the proposed approach, comprising seven steps. Step 1: we extract feature vectors from all samples in the target dataset for each backbone, perform distance calculation followed by neighbor-based refinement, and combine the distances across backbones. Step 2: our ensemble-based clustering algorithm is performed to propose pseudo-labels and filter outliers. Step 3: we update the learning rate following a warm-up strategy~\cite{luo2019strong}. Step 4: for each cluster obtained in Step 2, we randomly select a sample as a proxy. Steps 5, 6, and 7: a set of batches are created to optimize the backbones, each backbone is independently optimized and momentum models are updated based on the backbones' weights using a Mean Teacher strategy~\cite{tarvainen2017mean}. The red flow is performed $\lfloor C/P \rfloor$ times, where $C$ is the number of clusters in the current iteration, and $P$ is the number of clusters per batch. The cluster proxies are redefined $K_{2}$ times (green flow), after the red flow. The blue flow (entire pipeline) is performed $K_{1}$ times. Best viewed in color.}

\label{fig:overview_pipeline}
\end{figure*}

The training pipeline is composed of seven steps: feature extraction and neighborhood-based distance computation, ensemble-based clustering, learning rate update, proxy selection, batch creation, optimization, and Mean Teacher averaging. Figure~\ref{fig:overview_pipeline} depicts an overview of these steps.

In the first step, features are extracted for each sample, considering different backbones. We compute pairwise distances between samples based on their neighborhood for each backbone, and average them across backbones to obtain a more refined and unique distance matrix. The second step is the application of our ensemble-based clustering technique to obtain pseudo-labels for the samples. We perform the learning rate calculation in the third step. In the fourth step, a proxy feature vector is selected for each cluster and, in the fifth step, sample batches are created. This information is used during the sixth step, which is the optimization of each backbone, independently. In the last step, a Mean Teacher technique is used to combine the weights of a backbone over training steps in a \textit{momentum} model, which is used later for inference.

We use the term \textit{iteration} to refer to one complete iteration of the pipeline (blue flow in Figure~\ref{fig:overview_pipeline}), and \textit{epoch} to refer to when the proposed clusters are used for optimization in the current iteration (green flow in Figure~\ref{fig:overview_pipeline}). We perform $K_1$ iterations and $K_2$ epochs per iteration.

\subsection{Step 1: Feature extraction and neighborhood-based distance computation}

Consider a set $X = \{x_{i}\}^{N}_{i=1}$ of unlabeled data points in the target domain, consisting of $N$ samples; and $M$ backbones that generate feature representations for these samples. 

\begin{figure}[ht]
\centering
\includegraphics[width=1.9in]{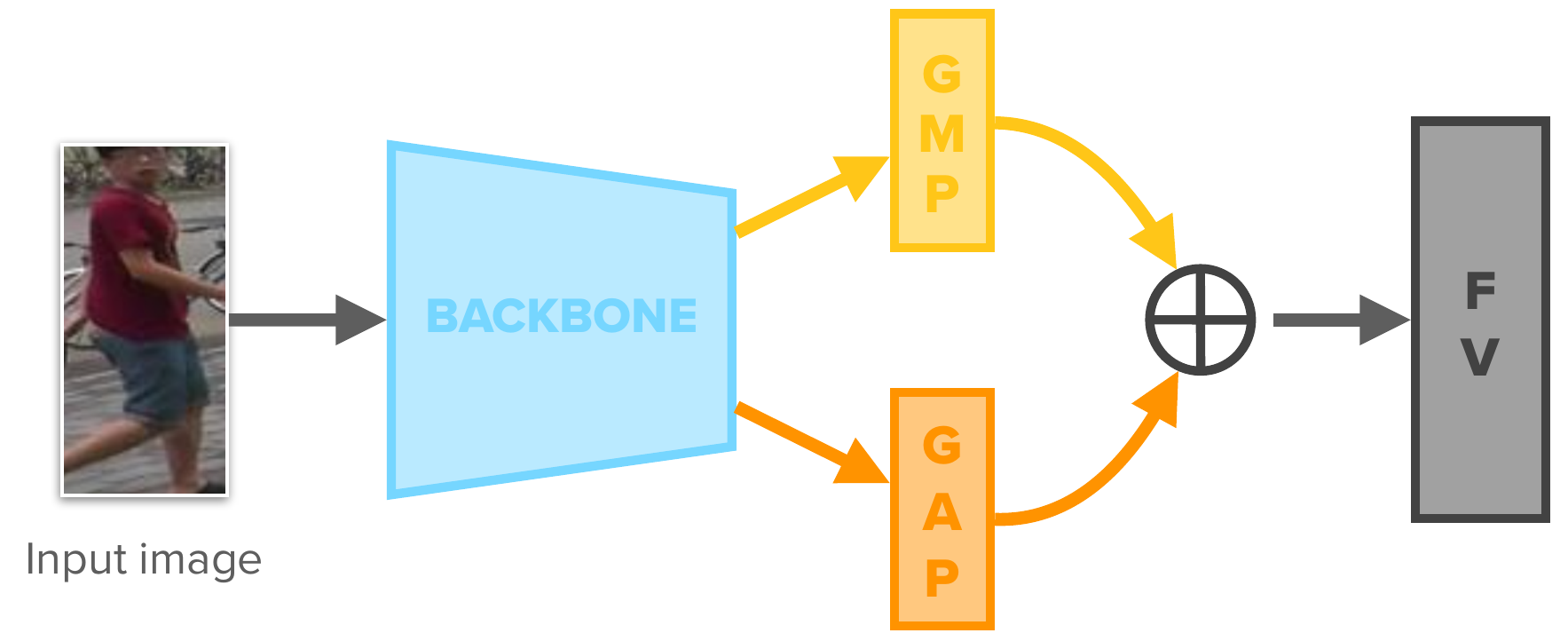}
\caption{The final feature vector (FV) is obtained by extracting both global max pooling (GMP) and global average pooling (GAP) of the previous layer's output and then adding the two resulting vectors element-wise.}
\label{fig:GAP+GMP}
\end{figure}

In prior art for image representation, the output of the last global max or average pooling layer is commonly used as the final feature vector. However, global max and average pooling operations produce distinct and complementary descriptions and, when used together, they can increase the quality of the representation~\cite{chen2020enhancing}. Following this idea, we perform both global max and average pooling after the last feature map and add the resulting vectors element-wise for the final feature vector (Figure~\ref{fig:GAP+GMP}). 
It is important to note that this is only done for images. For text representation, the output of the last layer is directly used as the final representation.

After extracting features for all samples, we L2-normalize them so that they are projected onto a unit hyper-sphere. Therefore, we have a set $F^{m} \in \mathbb{R}^{N \times d_{m}}$ of normalized feature vectors extracted by the $m$-th backbone, where $d_{m}$ represents the dimension of features in this set. 

For each $F^{m}$, we calculate pairwise distances for all samples in the set. Inspired by re-ranking techniques~\cite{zhong2017re}, we refine these distances by considering the neighborhood of the samples; i.e., we normalize the distance between two samples based exclusively on the number of neighbors in common. Each distance is a number between 0 (no neighbors in common) to 1 (all neighbors in common). Consequently, we have $M$ refined distance matrices $\{D_{1}, D_{2}, ..., D_{m}, ..., D_{M}\}$, one for each backbone. 
Distance matrix $D_{m}$ is a general representation of the knowledge obtained by the $m$-th backbone over the input data, as it is calculated based on the samples' feature vectors. To explore potential complementary knowledge, we propose averaging all distance matrices:

\begin{equation}
\label{eq:ensemble_matrices}
    \overline{\rm D} = \frac{1}{M}\sum_{m=1}^{M} D_{m}.
\end{equation}
The final distance matrix $\overline{\rm D}$ is used as input to the proposed ensemble-based clustering technique.

\subsection{Step 2: Ensemble-based Clustering}
\label{subsec:step2}

DBSCAN~\cite{ester1996density} clustering is the basis for our ensemble-based clustering. It relies on two hyper-parameters: $minPts$ -- the minimum number of samples on a point's neighborhood to consider it as a core point -- and $\varepsilon$ -- the size of the neighborhood. Two data points $p$ and $q$ are considered neighbors if the distance between them is less than $\varepsilon$.
A data point $p$ is a core point if it has at least $minPts$ neighbors. If it has less than $minPts$ but is neighbor to a core point, then $p$ is a border point. Otherwise, it is considered an outlier. Two points $p$ and $q$ are within the same cluster if a path exists $P = \{p_{0}, p_{1}, ..., p_{n}\}$, where $\forall_{1 \leq i \leq n-1} \text{ } p_{i}$ is a core point, $p = p_{0}$ and $q = p_{n}$.

The performance impact of the two hyper-parameters has been studied~\cite{hou2016dsets, chen2020enhancing}, and the conclusion is that DBSCAN is more sensitive to $\varepsilon$ than to $minPts$. A wrong choice of $\varepsilon$ can substantially hinder the performance, requiring domain knowledge to select its optimal value. A dataset with high intra-class variability might yield non-convex and sparse clusters, rendering the intra-class data points far away from each other while inter-class samples are closer. To account for this, a higher $\varepsilon$ would be needed to group sparse samples in the same cluster. In turn, datasets with lower intra-class variability might require a lower $\varepsilon$, as a larger value could introduce false positives in the same cluster. 

The described problem is common to several Person ReID benchmarks. 
Market1501~\cite{zheng2015scalable} is a dataset that comprises 751 identities recorded from six different cameras in the training set, while MSMT17~\cite{wei2018person} has 1041 identities recorded from fifteen different cameras. 
Identities on MSMT17 have larger intra-class variability than on Market1501 as the number of different views of an identity is prone to be higher. 
Thus, different datasets require different values of $\varepsilon$, and this has also been pointed out in prior art. In~\cite{Chen_2021_ICCV}, the authors use a lower value for Market1501 ($\varepsilon = 0.5$) and a larger value for MSMT17 ($\varepsilon = 0.6$) to account for dataset complexity. In~\cite{zhang2022implicit}, the authors adjust the value of $\varepsilon$ to obtain better results.
However, in a fully-unsupervised scenario, it is impossible to select an optimal value for $\varepsilon$ as there is no prior knowledge of the target data. Therefore, it is paramount to develop a clustering algorithm that does not depend on hyperparameter tuning.

We propose an ensemble-based clustering algorithm. As different values of $\varepsilon$ yield different clusters, we run DBSCAN with different $\varepsilon$ values and combine their results into a single final result. The proposed method effectively deals with noisy cases, allowing different but closer dense regions to be assigned to the same cluster, avoiding false positives and alleviating the burden of choosing the proper value for $\varepsilon$.

Considering the feature space defined by the refined distance matrix $\overline{\rm D}$ from Step 1, we perform DBSCAN with five $\varepsilon$ values: $0.5$, $0.55$, $0.6$, $0.65$, and $0.7$. As the neighborhood increases, more samples are assigned to a cluster. This does not mean that all samples are true positives and we need a way to detect false positives.
If a sample has been detected as an outlier with $\varepsilon = 0.5$, it is kept as an outlier on further runs.
We then can better filter out false positive samples while grouping closer dense true positive regions in the same cluster. 

\begin{figure*}[ht]
\centering
\includegraphics[width=5.9in]{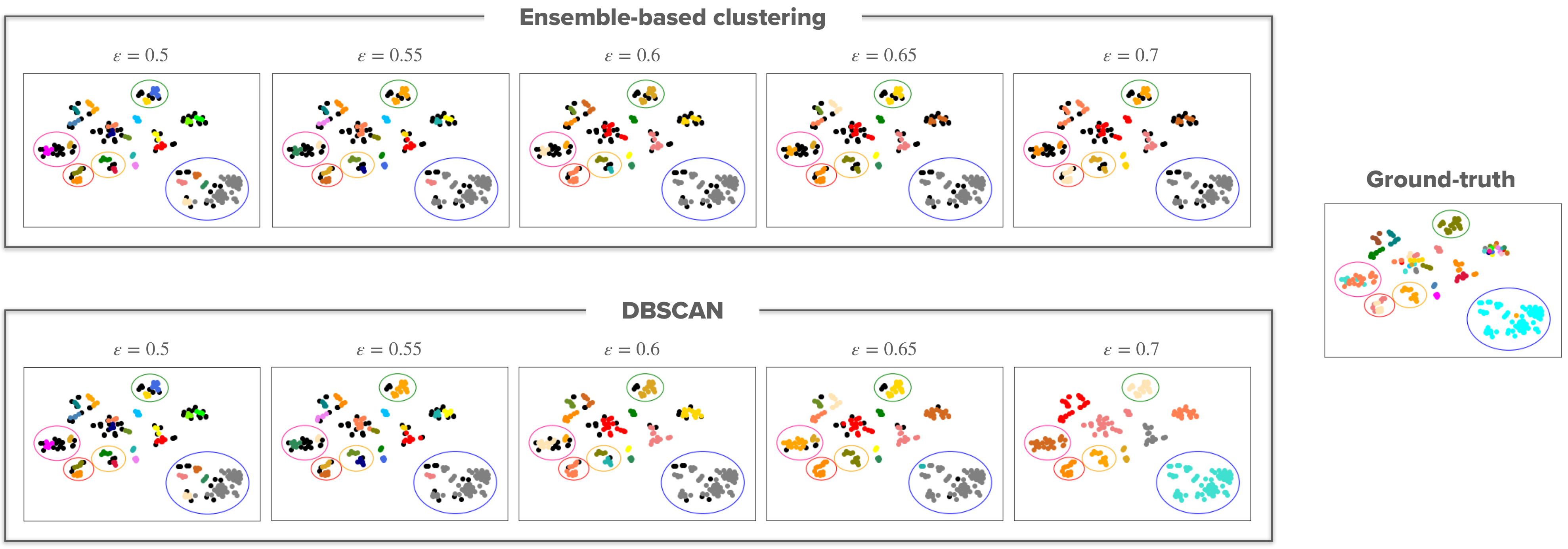}
\caption{Comparison between our proposed ensemble-based clustering method and DBSCAN. The data points are identities randomly sampled from the DukeMTMC-ReID dataset and projected into a 2D space by T-SNE~\cite{van2008visualizing}. Data points of the same color represent an identity. Black points are outliers. Circles highlight regions of interest to be analyzed. In our ensemble-based clustering, the results for an $\varepsilon$ value intrinsically combine results from all previous runs by keeping the data points' status if they were previously assigned as outliers or inliers. The last run ($\varepsilon = 0.7$) groups its results and all the previous ones, and produces more robust clusters than DBSCAN when we compare them to the final generated clusters with the ground truth. Best viewed in color.}
\label{fig:clustering_comparison}
\end{figure*}

Our ensemble-based clustering is illustrated and compared with the normal DBSCAN in Figure~\ref{fig:clustering_comparison}, where different colors represent different clusters (black represents outliers), and the circles highlight important regions to be analyzed. In the ground truth, the pink and red circles present noisy data with two identities mixed, which would be hard to split with a single $\varepsilon$ value. With $\varepsilon=0.5$, DBSCAN yields the highest true positive rate, but it cannot group sparser identities such as the one within the green circle, which is subdivided into two clusters (blue and yellow). The same happens within the blue circle: most samples are from the same identity, but DBSCAN subdivides it into five clusters. In turn, our method is able to group results from lower $\varepsilon$ values and create more robust clusters, as shown in Figure~\ref{fig:clustering_comparison}, $\varepsilon=0.7$.

One optimization is the reduction of DBSCAN runs. One must remember that if a point is detected as an outlier for $\varepsilon = 0.5$, it is kept as an outlier in subsequent runs with larger values. The same happens if a sample is an inlier for $\varepsilon = 0.5$; it will be an inlier in further runs as the neighborhood always increases. However, inlier samples assigned to different clusters can be put together in the same cluster due to the increasing neighborhood. For this reason, we say that the results for the $j$-th run contain the results for the $i$-th run, with $i \leq j$, as samples that are inliers or outliers keep their status in subsequent runs. 
It becomes clear that it is enough to run DBSCAN with $\varepsilon = 0.5$ and $\varepsilon = 0.7$ only, as the last run implicitly combines all intermediate results.

The main reason for choosing DBSCAN as the clustering algorithm is that it has the same or better resource consumption compared to other methods that have been used in cluster-based Unsupervised Person Re-Identification. Some previous works have employed Agglomerative Clustering, such as BUC~\cite{lin2019bottom}, MV-ReID~\cite{yin2021multi}, and HCT~\cite{zeng2020hierarchical}. Other works considered K-Means, such as PUL~\cite{fan2018unsupervised}, MMT~\cite{ge2020mutual} MEB-Net~\cite{zhai2020multiple}. We justify our choice in terms of three aspects: memory consumption, time complexity, and performance. 
    
\textit{Memory consumption}: Since our method relies on the Re-Ranking technique to refine the distance between samples based on neighborhood similarity, it is necessary to store a $N \times N$ distance matrix, where $N$ is the number of data samples. This has memory complexity of $\mathcal{O}(N^2)$. All clustering methods require the full distance matrix, so all of them are upper-bounded by the same complexity.

\textit{Time Complexity}: The Agglomerative Clustering~\cite{WHITTINGHAM202181} with the single-linkage variant has a time complexity of $\mathcal{O}(N^2)$. The K-Means clustering is well-known by the time complexity of $\mathcal{O}(NKT)$, where $K$ is the pre-defined number of clusters and $T$ is the number of iterations. As reported by~\cite{7065640}, $T \propto N$ so its effective time complexity is $\mathcal{O}(N^2)$. The DBSCAN, on the other hand, as reported in the original paper~\cite{ester1996density}, has a complexity of $\mathcal{O}(NlogN)$ in an optimal implementation, which makes it more efficient than Agglomerative Clustering and K-Means.

\textit{Performance}: Agglomerative Clustering performance depends on the choice of the linkage criteria to link the closest clusters. It is also sensible to outliers, because a single point can bridge two unrelated clusters and create large clusters with uncorrelated points. K-Means does not detect outliers since it enforces each point to belong to a cluster. It is also highly sensitive to the pre-definition of the number of clusters $K$ and to the selection of the initial clusters centers~\cite{7065640}. DBSCAN is able to detect outliers automatically. It creates clusters based on the reachability of the points, not depending on the number of clusters \textit{a priori}. It is only sensitive to the selection of the $\varepsilon$ hyper-parameter, which is the main motivation behind the proposed ensemble-based clustering.

\subsection{Step 3: Learning rate update}

To update the learning rate, we consider a warmup strategy~\cite{luo2019strong}, which is effective mainly in the first training iterations, where the number of samples is lower than in further iterations. In the first iterations, there is still little data available for training, as the outliers detected by the clustering algorithm are in greater quantity and are discarded as noisy samples. In this context, the model is more prone to overfitting, and a lower learning rate can aid the training in such cases. 

The warmup strategy consists of starting the training process with a small learning rate value and gradually increasing it along the first iterations.
Based on~\cite{luo2019strong}, we define the learning rate at iteration $t$ as
\begin{equation}
\label{eq:lr_reliability}
lr_{t}=  
\begin{cases}
    lr_{base}*\frac{t}{10}, & \text{$t \leq 10$}, \\    
    lr_{base}, & \text{$10 < t \leq K_1$}, \\
\end{cases}
\end{equation}
\noindent where $lr_{base}$ is a base value for the learning rate usually set to $3.5e^{-4}$, and $K_{1} = 30$. 
The learning rate linearly increases in the first ten iterations and is constant for the remaining ones.

\subsection{Step 4: Proxy selection}
\label{sec:proposed_step4}

Once pseudo-labels are assigned to unlabeled data samples based on the clustering results, and the learning rate is adjusted, the backbones can be updated.

This process starts in Step 4, with the selection of cluster proxies, which are prototypes that represent the clusters. For each cluster, its proxy is the feature vector of a randomly selected sample within that cluster. 

Although clusters tend to become more reliable as more iterations are performed, there is still a chance they might contain false positive samples. Assuming that the majority of samples are true positives, we hypothesize that a random selection is more likely to return a robust proxy than, for instance, computing the mean vector from all samples (true and false positives). We verify the impact of selecting a random or mean proxy in Section~\ref{sec:ablation_study}.

\subsection{Step 5: Batch creation}
\label{sec:proposed_step5}

The next step is batch creation. We consider the PK strategy~\cite{hermans2017defense}, where we randomly select $P$ out of $C$ clusters generated in the current iteration, and $K$ samples per cluster, creating batches of size $P*K$. 

It is important that a cluster appears only once per epoch, exposing the optimization to more diversity. For this, we create $\lfloor \frac{C}{P}\rfloor$ batches, which can leave some clusters out of the current epoch because of the rounding. However, as we perform $K_2$ epochs per iteration, the clusters not used in the current epoch will likely be selected in the next epochs.

\subsection{Step 6: Optimization}
\label{sec:proposed_step6}

The created batches are forwarded to the backbones for optimization.
The loss function $L$ to be minimized is composed of two other loss functions: $L_{proxy}$ and $L_{hard}$.

For the $m$-th backbone, loss function $L_{proxy}$ is based on cluster proxies $p_{j}^{m}$, $1 \leq j \leq C$ and is defined as
\begin{equation}
\label{eq:proxy_loss}
    \displaystyle L_{proxy}(B;\theta_{m}) = -\frac{1}{|B|}\sum_{i=1}^{|B|}log\left[\frac{exp(f_{i}^{m}.p_{+}^{m}/\tau)}{\sum_{j=1}^{C}exp(f_{i}^{m}.p_{j}^{m}/\tau)}\right],
\end{equation}
\noindent where $B$ is the batch, $|B|$ is the batch size, $\theta_{m}$ are the weights of the $m$-th backbone in the current iteration, $f_{i}^{m}$ is the feature vector of the $i$-th sample in $B$ extracted by the $m$-th backbone, $p_{+}^{m}$ is the proxy of the same cluster as the $i$-th sample in $B$, and $\tau$ is a temperature hyper-parameter to regulate the sharpness of the distribution of distances from the $i$-th sample to all proxies. The rationale is to approximate each sample in the batch from its respective proxy and keep it apart from the other proxies. 

As hard sample mining has shown promising results in prior art~\cite{chen2020enhancing, Chen_2021_ICCV}, we consider it by utilizing a hard instance-based softmax-triplet loss defined as

\begin{equation}
\label{eq:hard_loss}
    \begin{aligned}
    \displaystyle &L_{hard}(B;\theta_{m}) = \\
    &-\frac{1}{|B|}\sum_{i=1}^{|B|}log\left[\frac{exp(f_{i}^{m}.f_{+}^{m}/\tau)}{exp(f_{i}^{m}.f_{+}^{m}/\tau) + exp(f_{i}^{m}.f_{-}^{m}/\tau)}\right], 
    \end{aligned}
\end{equation}
\noindent where $f_{+}^{m}$ is the hardest positive sample in comparison to the $i$-th sample in $B$, i.e., it is the most distant feature from $f_{i}^{m}$ within the same cluster in the current batch. Analogously, $f_{-}^{m}$ is the hardest negative sample in comparison to $f_{i}^{m}$ in the current batch, i.e., it is the sample closest to $f_{i}^{m}$ but from another cluster.

The $L_{hard}$ loss provides a local view as it only considers samples in the current batch, while $L_{proxy}$ is a global loss since it considers all proxies from all clusters. The final loss combines them into a single function:
\begin{equation}
\label{eq:final_loss}
    L(B;\theta_{m}) = L_{proxy}(B;\theta_{m}) + \lambda L_{hard}(B;\theta_{m}),
\end{equation}
\noindent where $\lambda$ is a hyper-parameter to control the impact of $L_{hard}$. We provide the sensitivity of our method to $\tau$ and $\lambda$ values in Section~\ref{sec:ablation_study}.

Each backbone is trained independently but with the same set of pseudo-labels and learning rate obtained in previous steps. That is, the green flow in Figure~\ref{fig:overview_pipeline} is performed for each of the $M$ backbones, once at a time.

\subsection{Step 7: Mean Teacher average}

In this step, we leverage the Mean Teacher strategy~\cite{tarvainen2017mean}, which averages model weights over training steps to produce a final, more accurate model. It computes a teacher model as the average of consecutive student models.

For each backbone (student model), we keep a teacher (or momentum) model with the same architecture. After each optimization step, the weights of a backbone are used to update the respective momentum model by Exponential Moving Average:
\begin{equation}
\label{eq:ema_weights}
    \Theta_{m}^{(t)} := \beta\Theta_{m}^{(t-1)} + (1 - \beta)\theta_{m}^{(t)},
\end{equation}
\noindent where $\beta$ controls the inertia of the momentum weights over training, $t$ represents the iteration, and $\Theta_{m}$ and $\theta_{m}$ are the weights of the momentum and student models, respectively, that correspond to the $m$-th backbone.

\subsection{Inference}

After the training pipeline, inference is done by ranking all gallery samples based on the distance to a query sample. We extract feature vectors for all gallery and query sets using the momentum models from each backbone, which we denote as $F_{q}^{m}$ and $F_{g}^{m}$, respectively, with $1 \leq m \leq M$. For each $m$, we calculate pairwise distances between samples of $F_{q}^{m}$ and $F_{g}^{m}$, resulting in a distance matrix $D_{q2g}^{m} \in \mathbb{R}^{|Q| \times |G|}$, where $|Q|$ and $|G|$ are the number of samples in the query and gallery sets. 

A final distance matrix $\overline{\rm D}_{q2g}$ is obtained by averaging all matrices element-wise:
\begin{equation}
\label{eq:distance_ensemble_evaluation}
    \overline{\rm D}_{q2g} = \frac{1}{M}\sum_{m=1}^{M}D_{q2g}^{m}.
\end{equation}

Each row of $\overline{\rm D}_{q2g}$ holds the distances from a query to the gallery samples. 
We sort these distances to infer the closest class to the query sample.

\section{Experiments}

We perform experiments to validate our self-supervised learning pipeline, considering two applications: Person ReID and Authorship Attribution from short text messages.

\subsection{Datasets}
\label{subsec:datasets}
For \textbf{Person ReID}, we use three well-known large-scale datasets:

\begin{itemize}
    \item Market1501~\cite{zheng2015scalable}: 12,936 images of 751 identities in the training set and 19,732 images in the test set. The test set is divided into 3,368 images for the query set and 15,913 images for the gallery set. We removed ``junk'' images from the gallery set as done by all previous works, so 451 images were discarded. It has six non-overlapping cameras; each identity is captured by at least two.
    
    \item DukeMTMC-ReID~\cite{ristani2016performance}: 16,522 images of 702 identities in the training set and 19,889 images in the test set. The test set is divided into 2,228 query images and 17,661 gallery images of 702 other identities. It has eight cameras, and each identity is captured by at least two.

    \item MSMT17~\cite{wei2018person}: 32,621 images of 1,401 identities in the training set and 93,820 images of 3,060 identities in the test set. The test set is divided into 11,659 images for the query set and 82,161 images for the gallery. It has 15 cameras recording three day periods (morning, afternoon, and night) on four different days. Out of the 15 cameras, 12 are outdoor, and 3 are indoor. Each identity is captured by at least two cameras. It is the most challenging dataset.
\end{itemize}

As done in previous ReID works, we remove images from the gallery with the same identity and camera of the query to assess performance in a true cross-camera scenario.
For evaluation, we calculate the Cumulative Matching Curve (CMC), from which we report Rank-1 (R1), Rank-5 (R5), Rank-10 (R10), and mean Average Precision (mAP).

For \textbf{Authorship Attribution}, we adopt two subsets of a dataset of tweets~\cite{theophilo2021authorship}. The first subset contains messages from $100$ authors, which we randomly split into two sets: $50$ authors for training and $50$ for testing. In the training set, there are $400$ tweets per author. We divide the test set into query and gallery sets, with $20$ tweets per author in the query set and $300$ tweets per author in the gallery set. There are $1000$ authors in the second subset, with $500$ for training and another $500$ for testing. We select $70$ tweets per author for the training and gallery sets and $20$ tweets per author for the query. The rationale is to verify the capacity for generalization in two scenarios with different complexities. The smallest subset comprises fewer identities and more tweets per author; the other has $10$ times more authors and fewer tweets per author. Following the setup of the ReID validation, we keep disjoint authors for training and testing to verify the generalization capacity of the model. For evaluation, we compute mAP, R1, R5, and R10.

\subsection{Implementation Details}

We adopt $M=3$ backbones. For person ReID, the backbones are well-known Deep Convolutional Neural Network (DCNN) architectures: ResNet50~\cite{he2016deep}, OSNet~\cite{zhou2019omni}, and DenseNet121~\cite{huang2017densely}, all of them previously trained over the ImageNet dataset~\cite{deng2009imagenet}. For Authorship Attribution, we consider BERT~\cite{devlin2018bert}, BERTweet~\cite{nguyen2020bertweet}, and T5~\cite{raffel2019exploring} architectures. 

For optimization, we consider the Adam~\cite{kingma2014adam} optimizer with weight decay $0.00035$. The learning rate is set following the behavior in Equation~\ref{eq:lr_reliability} with $lr_{base} = 0.00035$. We implement the neighborhood-based distance with re-ranking~\cite{zhong2017re}, which relies on two parameters: $k_{1}$ which defines the k-reciprocal neighborhood size, and $k_{2}$, which defines the neighborhood size to average the distance representation. Following prior art, we set them to $k_{1} = 30$ and $k_{2} = 6$. For batch creation using the PK technique, we set $P = 16$ and $K=12$, totaling $192$ samples per batch.

The values for $\varepsilon$ in the proposed ensemble-based clustering are $0.5$, $0.55$, $0.6$, $0.65$, and $0.7$. We keep $minPts = 4$ in all DBSCAN runs as done in prior art. For the loss function, we set $\tau = 0.04$ and $\lambda = 0.5$ in Equations~\ref{eq:proxy_loss}, \ref{eq:hard_loss}, and \ref{eq:final_loss}. We analyze the sensitivity of our method to $\tau$ and $\lambda$ in Section~\ref{sec:ablation_study}.

The pipeline (blue flow in Figure~\ref{fig:overview_pipeline}) is executed for $K_{1} = 30$ iterations and the green flow is executed for $K_{2} = 7$ epochs, for each set of proposed clusters, and $\beta=0.999$ in Equation~\ref{eq:ema_weights}. 

The training pipeline is implemented using PyTorch~\cite{NEURIPS2019_9015}. The evaluation part and the OSNet backbone are implemented on Torchreid~\cite{torchreid}. We perform all experiments on three RTX5000 GPUs, each with 16 GB of RAM. One is used to perform re-ranking while the other is used to execute the whole training pipeline. The code is available at \url{https://github.com/Gabrielcb/Leveraging\_ensembles\_and\_self\_supervised\_fully\_ReID}.

\subsection{Person ReID}

We compare our pipeline applied to the Unsupervised Person Re-Identification problem with relevant methods in the literature. The results are shown in Table~\ref{tab:state_of_art_reid}, and the main pros and cons for each method are presented in Table~\ref{tab:comparison_prior_art}.

\begin{table*}[ht]
\caption{Comparison with relevant Person ReID methods considering three setups: camera-based, tracklet-based, and fully-unsupervised methods. Our work fits in the last category, which is the most challenging. We highlight the three best results in the fully-unsupervised scenario: the best one in \bv{blue}, the second best in \gv{green}, and the third in \ov{orange}. For the other categories, we only highlight the best result. Best viewed in color.}
\label{tab:state_of_art_reid}
\centering
\resizebox{1.0\textwidth}{!}{\begin{tabular}{|p{1.8cm}| p{1.5cm}|p{0.7cm}|p{0.7cm}|p{0.7cm}|p{0.7cm}|p{0.7cm}|p{0.7cm}|p{0.7cm}|p{0.7cm}|p{0.7cm}|p{0.7cm}|p{0.7cm}|p{0.7cm}|}
\hline
\multicolumn{1}{|c|}{} &
\multicolumn{1}{|c|}{} &
\multicolumn{4}{|c|}{\textbf{Market}} &
\multicolumn{4}{|c|}{\textbf{Duke}} &
\multicolumn{4}{|c|}{\textbf{MSMT17}} \\
\hline
Method & Reference & mAP & R1 & R5 & R10
& mAP & R1 & R5 & R10 & mAP & R1 & R5 & R10 \\ \hline
\multicolumn{14}{|c|}{\textit{Camera-based}} \\ \hline
SSL~\cite{lin2020unsupervised} & CVPR'20 & 37.8 & 71.7 & 83.8 & 87.4 & 28.6 & 52.5 & 63.5 & 68.9 & - & - & - & - \\
CCSE~\cite{lin2020unsupervisedccse} & TIP'20 & 38.0 & 73.7 & 84.0 & 87.9 & 30.6 & 56.1 & 66.7 & 71.5 & 9.9 & 31.4 & 41.4 & 45.7\\
MPRD~\cite{ji2021meta} & ICCV'21 & 51.1 & 83.0 & 91.3 & 93.6 & 43.7 & 67.4 & 78.7 & 81.8 & 14.6 & 37.7 & 51.3 & 57.1 \\ 
DSCE-MC~\cite{yang2021joint} & CVPR'21 & 61.7 & 83.9 & 92.3 & - & 53.8 & 73.8 & 84.2 & - & 15.5 & 35.2 & 48.3 & - \\
JVTC~\cite{li2020joint} & ECCV'20 & 47.5 & 79.5 & 89.2 & 91.9 & 50.7 & 74.6 & 82.9 & 85.3 & 17.3 & 43.1 & 53.8 & 59.4 \\ 
JGCL~\cite{chen2021joint} & CVPR'21 & 66.8 & 87.3 & 93.5 & 95.5 & 62.8 & 82.9 & 87.1 & 88.5 & 21.3 & 45.7 & 58.6 & 64.5 \\
IICS~\cite{xuan2021intra} & CVPR'21 & 72.9 & 89.5 & 95.2 & 97.0 & 64.4 & 80.0 & 89.0 & 91.6 & 26.9 & 56.4 & 68.8 & 73.4 \\
CAP~\cite{wang2020camera} & AAAI'21 & 79.2 & 91.4 & 96.3 & 97.7 & 67.3 & 81.1 & 89.3 & 91.8 & 36.9 & 67.4 & 78.0 & 81.4\\ 
CCTSE~\cite{9521886} & TIFS'21 & 67.7 & 89.5  & 94.8 & 96.5 & 68.8 & 82.4 & 90.6 & 92.5 & - & - & - & - \\
ICE~\cite{Chen_2021_ICCV} & ICCV'21 & 82.3 & 93.8 & 97.6 & 98.4 & \bv{69.9} & \bv{83.3} & \bv{91.5} & \bv{94.1} & 38.9 & 70.2 & 80.5 & 84.4 \\ 
PPLR~\cite{cho2022part} & CVPR'22 & \bv{84.4} & \bv{94.3} & \bv{97.8} & \bv{98.6} & - & - & - & - & \bv{42.2} & \bv{73.3} & \bv{83.5} & \bv{86.5} \\
\hline
\multicolumn{14}{|c|}{\textit{Tracklet-based}}  \\ \hline
Star-Dac~\cite{prasad2022spatio} & PR'21 & 33.9 & 67.0 & 80.6 & 84.9 & 31.6 & 56.4 & 72.1 & 76.5 & - & - & - & - \\
TAUDL~\cite{li2018unsupervised} & ECCV'18 & 41.2 & 63.7 & - & - & 43.5 & 61.7 & - & - & - & - & - & - \\
TSSL~\cite{wu2020tracklet} & AAAI'20 & 43.3 & 71.2 & - & - & 38.5 & 62.2 & - & - & - & - & - & - \\ 
UTAL~\cite{li2019unsupervised} & TPAMI'20 & 46.2 & 69.2 & - & - & 44.6 & 62.3 & - & - & 13.1 & 31.4 & - & - \\
CycAs~\cite{wang2020cycas} & ECCV'20 & 64.8 & 84.8 & - & - & \bv{60.1} & \bv{77.9} & - & - & \bv{26.7} & \bv{50.1} & - & - \\ 
UGA~\cite{wu2019unsupervised} & ICCV'19 & \bv{70.3} & \bv{87.2} & - & - & 53.3 & 75.0 & - & - & 21.7 & 49.5 & - & - \\ \hline
\hline
 \multicolumn{14}{|c|}{\textit{Fully Unsupervised}}  \\ \hline
BUC~\cite{lin2019bottom} & AAAI'19 & 38.3 & 66.2 & 79.6 & 84.5 & 27.5 & 47.4 & 62.6 & 68.4 & - & - & - & - \\
GPUFL~\cite{sun2021unsupervised} & ICIP'21 & 42.3 & 69.6 & - & - & 37.7 & 57.4 & - & - & - & - & - & -  \\
MV-ReID~\cite{yin2021multi} & SPL'21 & 45.6 & 73.3 & 85.3 & 89.1 & 31.7 & 54.5 & 67.5 & 72.1 & - & - & - & - \\
MMCL~\cite{wang2020unsupervised} & CVPR'20 & 45.5 & 80.3 & 89.4 & 92.3 & 40.2 & 65.2 & 75.9 & 80.0 & 11.2 & 35.4 & 44.8 & 49.8 \\ 
HCT~\cite{zeng2020hierarchical} & CVPR'20 & 56.4 & 80.0 & 91.6 & 95.2 & 50.7 & 69.6 & 83.4 & \ov{87.4} & - & - & - & - \\
ABMT~\cite{chen2020enhancing} & WACV'20 & 65.1 & 82.6 & - & - & 63.1 & 77.7 & - & - & - & - & - & - \\
SpCL~\cite{ge2020self} & NeurIPS'20 & 73.1 & 88.1 & 95.1 & 97.0 & - & - & - & - & 19.1 & 42.3 & 55.6 & 61.2 \\
RLCC~\cite{zhang2021refining} & CVPR'21 & 77.7 & 90.8 & 96.3 & 97.5 & \ov{69.2} & \gv{83.2} & \bv{91.6} & \bv{93.8} & 27.9 & 56.5 & 68.4 & 73.1 \\
ICE~\cite{Chen_2021_ICCV} & ICCV'21 & 79.5 & 92.0 & 97.0 & \ov{98.1} & 67.2 & 81.3 & 90.1 & \gv{93.0} & 29.8 & 59.0 & 71.7 & 77.0 \\
CACL~\cite{li2022cluster} & TIP'22 & 80.9 & 92.7 & \gv{97.4} & \gv{98.5} & \gv{69.6} & \ov{82.6} & \gv{91.2} & \bv{93.8} & 23.0 & 48.9 & 61.2 & 66.4 \\
PPLR~\cite{cho2022part} & CVPR'22 & \ov{81.5} & \ov{92.8} & \ov{97.1} & \ov{98.1} & - & - & - & - & \ov{31.4} & \ov{61.1} & \ov{73.4} & \ov{77.8} \\
ISE~\cite{zhang2022implicit} & CVPR'22 & \bv{84.7} & \bv{94.0} & \bv{97.8} & \bv{98.8} & - & - & - & - & \gv{35.0} & \gv{64.7} & \gv{75.5} & \gv{79.4} \\
\hline
\textbf{Ours} &  & \gv{83.4} & \gv{92.9} & \ov{97.1} & 97.8 & \bv{72.7} & \bv{83.9} & \ov{91.0} & \gv{93.0} & \bv{42.6} & \bv{68.2} & \bv{77.9} & \bv{81.4} \\ \hline

\hline
\end{tabular}}

\end{table*}

We outperform all methods in the fully-unsupervised setup in the most challenging datasets, Duke and MSMT17, and obtain the second-best result in Market dataset, regarding mAP and R1. More specifically, we outperform the recent CACL method by $3.1$ and $1.3$ percentage points (p.p.) in mAP and R1, respectively, in the Duke dataset; and outperform ISE by $7.6$ and $3.5$ p.p. on MSMT17, the most challenging unsupervised ReID dataset. In the Market dataset, our results are the second best, considering mAP and R1. As we designed our method to tackle general and complex fully-unlabeled scenarios for multi-modal tasks, it achieves the best results in the most challenging datasets. The other methods were designed specifically for Person ReID and, for this reason, they are usually better in less complex datasets such as Market. This shows the effectiveness of our method in the fully-unsupervised scenario.

Other works assume metadata, such as camera labels and tracklets, but no identity information. In Table~\ref{tab:state_of_art_reid}, we can see that the most helpful metadata is camera information. Person ReID is naturally a cross-camera retrieval task: a method must be able to retrieve (from the gallery) images of the same identity used as a query but seen from other cameras. In this sense, camera information provides a significant impact if it is leveraged during training. This is evinced when we compare our results with the camera-based method PPLR. It considers camera proxies per cluster and pulls images from different cameras closer to overcome differences brought by different points of view. This seems especially beneficial in more complex datasets (MSMT17), where PPLR outperforms our method by $5.1$ p.p. in R1. However, we still attain the best performance in mAP for this dataset among all methods in any category. The same conclusion is drawn for the Duke dataset where our method outperforms all prior art in all scenarios in terms of mAP and R1. 

Tracklet-based methods explore temporal information during training but are not as competitive as camera-based ones. We outperform the state-of-the-art methods in all datasets by a large margin. 

Finally, our method is able to place most true positive samples from the gallery closer to the query for most of the datasets, which is represented by the highest mAP for the two most difficult datasets, Duke and MSMT17. Our method is thus able to achieve the best performance among all methods even over those assuming \textbf{strong camera information as metadata.} One should also note that, in general, our assumptions are even more relaxed than other methods. Our clustering algorithm, for instance, does not require hyper-parameter tuning, while ICE explicitly fixes a value for $\varepsilon$ depending on the target dataset.

In the Supplementary Material, we also compare our pipeline to UDA methods that require a source domain to provide initial task-related knowledge. We outperform all these methods in mAP, and obtain the best or second-best ranking values considering other metrics without source domain and any kind of labels.

\subsubsection{Training time}

We also analyze our pipeline in terms of execution time for training (Table~\ref{tab:time_evaluation}). Clustering (Step 2) time is negligible due to its optimized implementation. The fine-tuning process (Steps 3 to 7) takes longer as it requires optimization using the generated clusters for $K_{2} = 7$ epochs. The total time for the pipeline is in the order of a few hours, and it depends on the size of the dataset. For MSMT17, the largest one with $32,621$ images in the training set and 1,041 identities (greater than the number of classes on ImageNet), the method presents a reasonable time of around 19 hours.

\begin{table}[ht]
\caption{Execution time. We report the average time taken by some steps and the total time to execute the pipeline, for the three ReID datasets. The time format is HH:MM:SS.}
\label{tab:time_evaluation}
\centering
\begin{tabular}{|p{1.0cm}|p{0.9cm}|p{0.9cm}|p{0.9cm}|p{0.9cm}| P{1.3cm}|}
\hline
Dataset & \textbf{Step 1} & \textbf{Step 2} & \textbf{Steps 3 to 7} & \textbf{Total Time} & \textbf{Inference Time (ms)} \\ \hline
Market & 00:01:29 & 00:00:07 & 00:08:27 & 05:10:35 & 41.4\\
Duke & 00:01:53 & 00:00:10 & 00:15:13 & 08:47:37 & 59.1 \\
MSMT17 & 00:04:25 & 00:00:38 & 00:30:12 & 19:06:05 & 80.2 \\
\hline
\end{tabular}
\end{table}

Inference time also increases with the size of the gallery sets, as there are more samples to compare to the query. Inference for a query on Market, the smallest one, takes 41.4 milliseconds, while for MSMT17, the largest dataset, it takes 80.2 milliseconds. All scenarios present reasonable inference time under one second. 


\subsubsection{Qualitative Analysis}
\begin{figure*}[ht]
\centering
\subfloat[Duke -- Successful case.]{\includegraphics[width=3.3in]{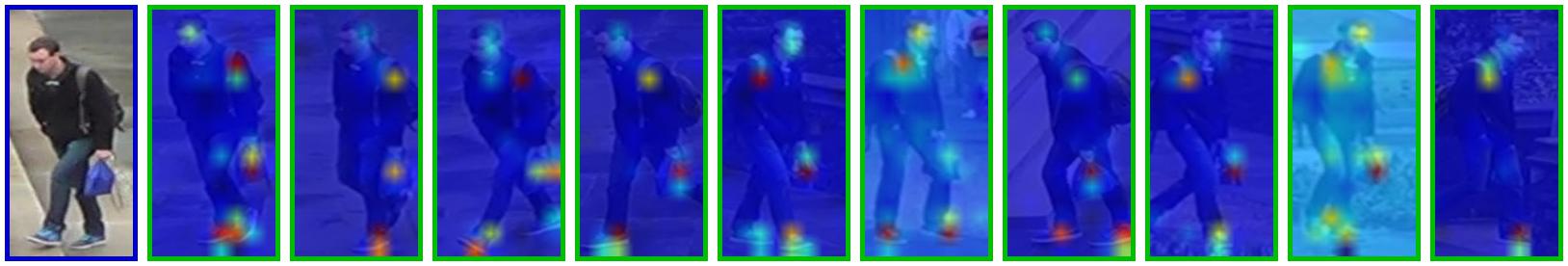}
\label{fig:Duke_successful}}
\hfil
\subfloat[Duke -- Failure case.]{\includegraphics[width=3.3in]{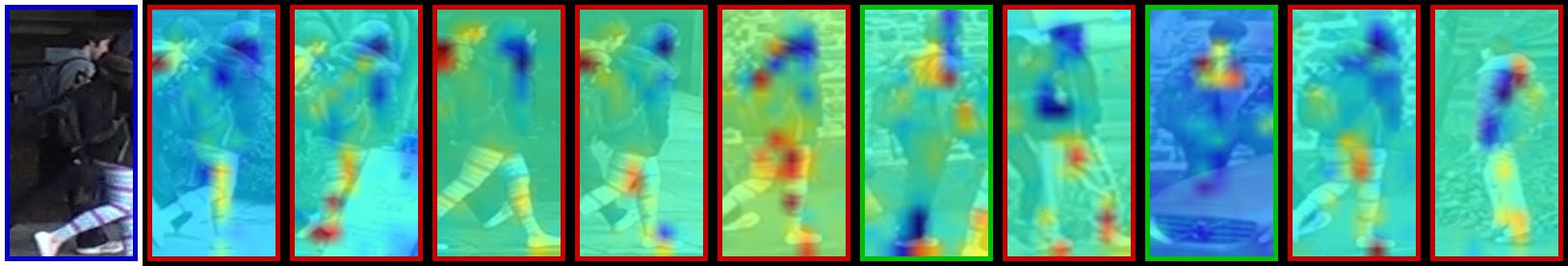} 
\label{fig:Duke_failure}}
\hfil
\subfloat[Market -- Successful case.]{\includegraphics[width=3.3in]{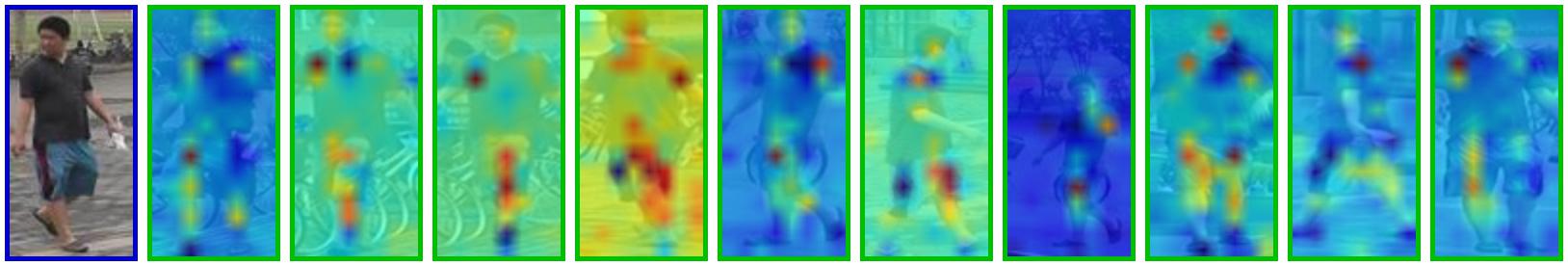} 
\label{fig:Market_successful}}
\hfil
\subfloat[Market -- Failure case.]{\includegraphics[width=3.3in]{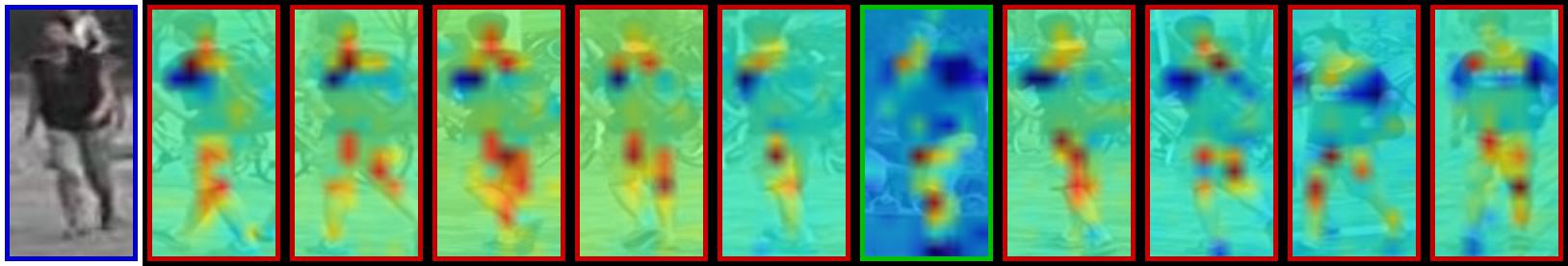} 
\label{fig:Market_failure}}
\caption{Success and failure examples for DukeMTMCReID and Market datasets. The images show the most activated regions (in red) in the gallery given a query image, considering the ResNet50 backbone. In each example, the query is the first image. Images with a green frame are correctly retrieved gallery images, and a red frame indicates a different identity from the query. 
}
\label{fig:qualitative_analysis_reid}
\end{figure*}

In terms of qualitative results, we provide some examples of success and some for failure, for DukeMTMCReID and Market datasets, illustrated in Figure~\ref{fig:qualitative_analysis_reid}. We do not show examples for MSMT17 as reproducing this dataset's images is not allowed in any format.

From the successful examples (Figures~\ref{fig:Duke_successful} and~\ref{fig:Market_successful}), we observe that our method can identify several fine-grained details throughout the image in order to retrieve the correct identity. This means it overcomes differences in point-of-view, pose, illumination, and background. In one of the failure examples (Figure~\ref{fig:Duke_failure}), there are two people in the same bounding box, which is ambiguous. In the second failure example (Figure \ref{fig:Market_failure}), the query presents an identity in low resolution and with clothing colors similar to the background. Despite being failure cases, the model can still find similar and fine-grained details, which could help in cases where the goal is to find people dressing similarly. We present more success and failure cases in the Supplementary Material.

\subsection{Authorship Attribution}

We now consider a second task --- Authorship Attribution --- with minor adjustments mainly related to the nature of the problem. For the backbones, we consider BERT~\cite{devlin2018bert}, BERTweet~\cite{nguyen2020bertweet}, and T5~\cite{raffel2019exploring} as they were developed to deal with text. We apply augmentation on tweets with more than $5$ tokens by masking from $10\%$ to $20\%$ of the tokens with a ``mask'' token on BERT and BERTweet, and with an ``unknown'' token on T5. Even the base version of BERT is too complex for short signals (text tweets), making the training more prone to overfitting. Hence, we freeze the first ten attention blocks of BERT and BERTweet, leaving only the $11^{th}$ block to be updated. For the same reason, we set $K_{1} = 15$ as the number of iterations --- half of the value used for the ReID experiments --- to alleviate the impact of over-training. We set $P = 8$ for batch creation and $K = 8$. All other parameters are the same as those used for ReID. The only difference in the pipeline is that we do not apply the optimization shown in Figure~\ref{fig:GAP+GMP}, as it is only for image representations.

We run comparative experiments considering the AdHominem method~\cite{boenninghoff2019explainable}. It employs an attention-based model for Authorship Attribution on social media text. AdHominem performs supervised training using Siamese networks to answer whether or not the same author wrote two tweets. To perform the raking task, we take the Euclidean distance between two tweets returned by the model and rank the gallery tweets given a query. The results are shown in Table~\ref{tab:authorship_performance}.

\begin{table}[ht]
\caption{Results for the Authorship Attribution task for two subsets of tweets: one with $50$ authors on training and $50$ authors on test, and the other with $500$ authors on training and $500$ authors on test. The best results are in \bv{blue}.}
\label{tab:authorship_performance}
\centering
\begin{tabular}{|p{1.85cm}|p{0.45cm}|p{0.4cm}|p{0.4cm}|p{0.4cm}|p{0.45cm}|p{0.4cm}|p{0.4cm}|p{0.4cm}|}
\hline
\multicolumn{1}{|c|}{} &
\multicolumn{4}{|c|}{\textbf{1\textsuperscript{st} subset} ($50$ authors)} &
\multicolumn{4}{|c|}{\textbf{2\textsuperscript{nd} subset} ($500$ authors)} \\
\hline
Model & mAP & R1 & R5 & R10 & mAP & R1 & R5 & R10\\ \hline
AdHominem~\cite{boenninghoff2019explainable} & 7.3 & 25.5 & 50.9 & 61.5 & 2.4 & 10.8 & 23.5 & 31.6 \\ \hline
\textbf{Ours} & \bv{14.3} & \bv{50.0} & \bv{73.0} & \bv{80.3} & \bv{5.0} & \bv{22.5} & \bv{37.0} & \bv{44.8}  \\ \hline
\end{tabular}
\end{table}

Our method outperforms AdHominem in both subsets. More specifically, we outperform AdHominem by $7.0$ and $24.5$ p.p. in mAP and R1, respectively, in the first subset. In the second one, we outperform it by $2.6$ and $11.7$ p.p.

One must note that AdHominem is trained in a \textbf{supervised} manner considering the identity of each tweet, i.e., the method knows \textit{a priori} ``who'' wrote the tweets to supervise the training. However, our method relaxes this constraint by taking only the raw tweet text \textbf{without any labeling.} Moreover, our training and test data are disjoint on the identities, and since AdHominem is trained for a closed set of authors, it generalizes poorly to unseen authors.

Although our method utilizes pre-initialized weights, these weights were trained for other tasks, such as question answering, next sequence prediction, predicting missing words, and so on, instead of Authorship Attribution.

\begin{figure}[ht]
\centering
\subfloat[50 authors.]{\includegraphics[width=1.5in]{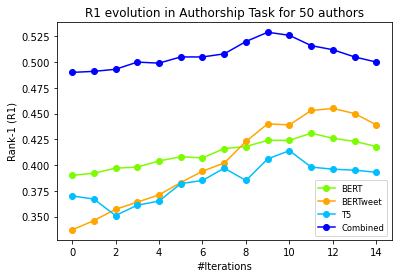}
\label{fig:authorship_50authros}}
\hfil
\subfloat[500 authors.]{\includegraphics[width=1.5in]{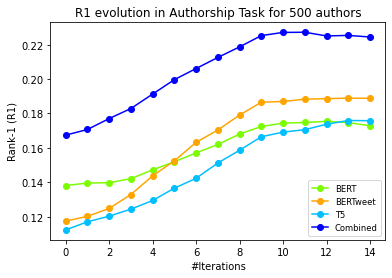}
\label{fig:authorship_500authors}}
\hfil
\caption{Rank-1 evolution over training iterations for the Authorship Attribution task considering a test set with (a) 50 authors and (b) 500 authors. ``Combined'' is obtained by averaging the final distances acquired with each backbone.}
\label{fig:authorship_performance}
\end{figure}

In Figure~\ref{fig:authorship_performance}, we provide the evolution of Rank-1 during training to show the merits of employing our pipeline. In the first subset ($50$ authors), we verify a Rank-1 oscillation after the $10^{th}$ iteration and a slight decrease until the last iteration. BERTweet provides the greatest gain, followed by BERT and T5. In the second subset ($500$ authors), the training process is more stable though it is numerically inferior as there are more authors. The feature space is denser, securing more stability for the convergence of the models. 

\subsection{Further Analysis}

Our applications are Unsupervised Person Re-Identification and Unsupervised Text Authorship Attribution, which are related to event analysis in a forensic context. Thus, it is important to understand how the performance of our approach is impacted when we consider common forensics features used for other tasks, such as image manipulation detection~\cite{barni2020improving}, or pre-processed inputs used in the GAN-generated image detection task~\cite{barni2020cnn}.

To test this, we apply the method proposed in~\cite{barni2020improving} for manipulation detection to the Unsupervised Person Re-Identification problem. It is an ensemble-based algorithm with forensics features.
We employ the same features---SPAM~\cite{5437325} and CRSPAM1372~\cite{barni2018detection}---, which have also shown to achieve top-tier performance on the manipulation detection task.

However, since Unsupervised Person Re-Identification differs in nature from manipulation detection, SPAM and CRSPAM1372 features induce a huge performance drop when employed in our setup. These hand-crafted features have not been designed to overcome the non-linearity in Unsupervised Person Re-Identification problems caused by the difference in illumination, resolution, pose, background, and occlusion.

We also consider pre-processing techniques employed in the GAN-generated image detection task. We take the method proposed in~\cite{barni2020cnn}, whose main goal is distinguishing between GAN-generated and real images. The authors argue that it is harder for GANs to reconstruct consistent relationships among the color bands, and they propose an approach to detect these inconsistencies.


For a fair comparison, we keep our whole pipeline and change only the input of our backbones to match the ones from~\cite{barni2020cnn}. Their method induces a huge performance drop compared to ours with standard RGB images. Our performance is better than theirs in all metrics in all evaluated datasets. Their method requires a signal analysis across color bands to reconstruct inconsistencies and to detect the neighboring inconsistencies and noise introduced by GAN-based models. However, this process might destroy the semantics present in the images, which is fundamental in Unsupervised Person Re-Identification. Our models, conversely, learn high-level semantic and camera-invariant features to match the same person seen from different camera views and distinguish from other people by looking at discriminant parts, such as clothes, shoes, bags, and faces. (Figure~\ref{fig:qualitative_analysis_reid}). In summary, the method proposed in~\cite{barni2020cnn} looks for fine-grained details at the noise level, while ours looks for fine-grained details at the semantic level. Further details are provided in the Supplementary Material.

\section{Ablation Study}
\label{sec:ablation_study}

We validate each part of our pipeline by checking how these influence the performance, considering the Market and Duke datasets. If not specified, we assume ensemble-based clustering, and $\tau = 0.04$ and $\lambda = 0.5$ in Equations~\ref{eq:proxy_loss},~\ref{eq:hard_loss}, and~\ref{eq:final_loss}.

\subsection{Step 1: impact of distance averaging}

In the first step of the pipeline, we average the distance matrices obtained with each backbone separately, computing a combined distance matrix $\overline{\rm D}$ (Equation~\ref{eq:ensemble_matrices}). This allows complementary knowledge to be grouped together for training. To measure how this impacts the final performance, we train each backbone separately, as expected, but feed each distance matrix directly to Step 2, instead of averaging them together. We present the results in Table~\ref{tab:ablation_ensemble}, showing the impact of averaging the distances in Step 1 for each backbone separately and for the combined result (considering Equation~\ref{eq:distance_ensemble_evaluation}).


\begin{table}[ht]
\caption{Impact when we remove the proposed backbones knowledge combination (Equation~\ref{eq:ensemble_matrices}). Results with (*) mean we do not apply our proposed fusion. Best results are in \bv{blue}.}
\label{tab:ablation_ensemble}
\centering
\begin{tabular}{|p{1.7cm}|p{0.4cm}|p{0.4cm}|p{0.4cm}|p{0.4cm}|p{0.4cm}|p{0.4cm}|p{0.4cm}|p{0.4cm}|}
\hline
\multicolumn{1}{|c|}{} &
\multicolumn{4}{|c|}{\textbf{Market}} &
\multicolumn{4}{|c|}{\textbf{Duke}} \\
\hline
Model & mAP & R1 & R5 & R10 & mAP & R1 & R5 & R10\\ \hline
ResNet50* & 77.8 & 90.7 & 96.2 & 97.7 & 65.6 & 80.0 & 89.2 & 91.3 \\
ResNet50 & \bv{80.4} & \bv{91.1} & \bv{96.8} & \bv{98.0} & \bv{69.4} & \bv{82.0} & \bv{90.4} & \bv{92.9} \\ \hline
OSNet* & 73.2 & 87.8 & 95.2 & 96.8 & 67.5 & 81.3 & 90.0 & \bv{92.6} \\
OSNet & \bv{78.6} & \bv{90.5} & \bv{96.1} & \bv{97.2} & \bv{68.9} & \bv{82.9} & \bv{90.2} &  92.0\\ \hline 
DenseNet121* & 73.2 & 87.5 & 94.9 & 96.6 & 63.8 & 79.2 & 87.8 & 90.4 \\
DenseNet121 & \bv{78.9} & \bv{90.3} & \bv{95.9} & \bv{97.4} & \bv{67.7} & \bv{82.1} & \bv{90.0} & \bv{92.2} \\ \hline
Ours* & 81.0 & 91.8 & 96.9 & \bv{97.9} & 70.6 & 82.3 & 90.2 & 92.5 \\
Ours & \bv{83.4} & \bv{92.9} & \bv{97.1} & 97.8 & \bv{72.7} & \bv{83.9} & \bv{91.0} & \bv{93.0} \\

\hline
\end{tabular}
\end{table}

Distance averaging in Step 1 is important for training, and it positively impacts each backbone individually, as well as their final combination, which allows for better grouping in the clustering step. Considering the combination of backbones (Equation~\ref{eq:distance_ensemble_evaluation}), the gains are also considerable for both datasets. In Market, we achieve an improvement of $2.4$ p.p. and $1.1$ p.p., in mAP and R1, respectively, and $2.1$ p.p. and $1.6$ p.p. on Duke. These results show the effectiveness of our proposed approach without requiring mutual training~\cite{ge2020mutual, zhai2020multiple} or co-teaching~\cite{yang2020asymmetric}, which in turn promotes simpler training. Moreover, from Table~\ref{tab:ablation_ensemble}, it is possible to conclude that different backbones provide complementary information, as there is an improvement for all metrics when they are combined (Equation~\ref{eq:distance_ensemble_evaluation}), for both setups (``Ours(*)'' and ``Ours'').

\subsection{Step 2: impact of the ensemble-based clustering}

We verify the effectiveness of our proposed ensemble-based clustering method. We replace it with the standard DBSCAN algorithm in the second step of our pipeline, and keep the remaining parts unchanged. As we combine the results of DBSCAN runs with different $\varepsilon$ values into a single result, in Table~\ref{tab:ablation_clustering}, we present the separate results for each $\varepsilon$ value.

\begin{table}[ht]
\caption{Impact of the ensemble-based clustering. We replace it with the standard DBSCAN by fixing five $\varepsilon$ values. The best result for each metric is highlighted in \bv{blue}.}
\label{tab:ablation_clustering}
\centering
\begin{tabular}{|p{1.5cm}|p{0.4cm}|p{0.4cm}|p{0.4cm}|p{0.4cm}|p{0.4cm}|p{0.4cm}|p{0.4cm}|p{0.4cm}|}
\hline
\multicolumn{1}{|c|}{} &
\multicolumn{4}{|c|}{\textbf{Market}} &
\multicolumn{4}{|c|}{\textbf{Duke}} \\
\hline
DBSCAN $\varepsilon$ & mAP & R1 & R5 & R10 & mAP & R1 & R5 & R10\\ \hline
0.50 & 81.6 & 90.7 & 95.4 & 96.4 & 60.9 & 76.4 & 82.7 & 84.7 \\
0.55 & 82.9 & 92.3 & 96.4 & 97.3 & 62.8 & 77.2 & 83.7 & 85.8 \\
0.60 & 82.3 & 91.6 & 96.2 & 97.4 & 67.5 & 79.6 & 87.7 & 89.9 \\
0.65 & 82.7 & 92.3 & 96.7 & 97.7 & 69.1 & 81.9 & 88.9 & 91.5 \\
0.70 & 81.8 & 91.8 & 96.8 & \bv{97.8} & 70.3 & 82.8 & 90.0 & 92.1 \\ \hline
\textbf{Ours} & \bv{83.4} & \bv{92.9} & \bv{97.1} & \bv{97.8} & \bv{72.7} & \bv{83.9} & \bv{91.0} & \bv{93.0}  \\ \hline
\end{tabular}
\end{table}

This experiment shows that the proposed ensemble-based clustering effectively combines DBSCAN intermediate results into a final one, suggesting more robust clusters, and outperforming all results in both datasets. Note that, as motivated in Section~\ref{subsec:step2}, if a single $\varepsilon$ is employed for clustering, each dataset has different optimal values. In Table~\ref{tab:ablation_clustering}, we see that the model achieves the best performance in mAP and R1 with $\varepsilon = 0.55$ for Market and $\varepsilon = 0.7$ for Duke, which shows that the optimal values can change significantly from a dataset to another. In contrast, our proposed ensemble-based clustering strategy obtains the best performance by grouping DBSCAN results using lower $\varepsilon$ values (denser clusters, lower false positive rate) and higher $\varepsilon$ values (more diverse clusters, lower false negative rate), alleviating the burden of choosing a proper unique value for this hyper-parameter.

\subsection{Step 3: impact of proxy selection}
In Step 4, we select a random sample per cluster as a proxy to aid the optimization as in~\cite{dai2021cluster}. We validate this choice by replacing the random selection with the mean feature vector of each cluster. The results are shown in Table~\ref{tab:proxy_ablation}. 

\begin{table}[ht]
\caption{Impact of the proxy selection. We replace the random selection of samples to serve as cluster proxies by the mean feature vector of the cluster. The best results are in \bv{blue}.}
\label{tab:proxy_ablation}
\centering
\begin{tabular}{|p{1.8cm}|p{0.4cm}|p{0.4cm}|p{0.4cm}|p{0.4cm}|p{0.4cm}|p{0.4cm}|p{0.4cm}|p{0.4cm}|}
\hline
\multicolumn{1}{|c|}{} &
\multicolumn{4}{|c|}{\textbf{Market}} &
\multicolumn{4}{|c|}{\textbf{Duke}} \\
\hline
& mAP & R1 & R5 & R10 & mAP & R1 & R5 & R10\\ \hline
Mean & 82.0 & 91.8 & 96.2 & 97.5 & 70.7 & 81.8 & 90.0 & 92.2 \\
\textbf{Ours} (random) & \bv{83.4} & \bv{92.9} & \bv{97.1} & \bv{97.8} & \bv{72.7} & \bv{83.9} & \bv{91.0}  & \bv{93.0} \\
\hline
\end{tabular}
\end{table}

Random selection improves the performance for all metrics, but mainly for mAP and R1. More specifically, we obtain a gain of $1.4$ and $1.1$ p.p. in mAP and R1, respectively, in the Market dataset, and $2.0$ and $2.1$ p.p. in the Duke dataset. This validates our assumption that using a mean vector as a proxy hinders cluster representation and further training, as it is affected by the false positive samples of the cluster. 

\subsection{Step 4: impact of loss function hyper-parameters}

We vary hyper-parameters $\tau$ and $\lambda$ in the loss functions (Equations~\ref{eq:proxy_loss},~\ref{eq:hard_loss}, and~\ref{eq:final_loss}) to check how they impact the pipeline in both image- and text-based applications.

\begin{figure}[ht]
\centering
\subfloat[Different $\tau$ values on Market.]{\includegraphics[width=1.5in]{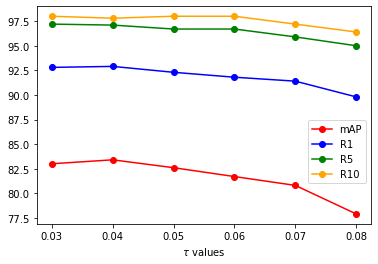}
\label{fig:tau_market}}
\hfil
\subfloat[Different $\tau$ values on Duke.]{\includegraphics[width=1.5in]{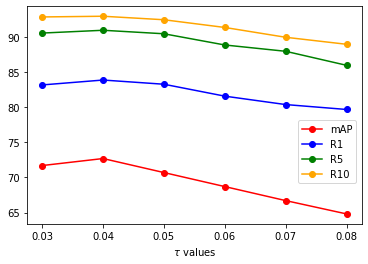}
\label{fig:tau_duke}}
\hfil
\subfloat[Different $\tau$ values on 50 authors.]{\includegraphics[width=1.5in]{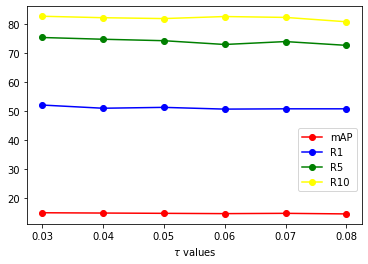}
\label{fig:tau_50authors}}
\hfil
\subfloat[Different $\tau$ values on 500 authors.]{\includegraphics[width=1.5in]{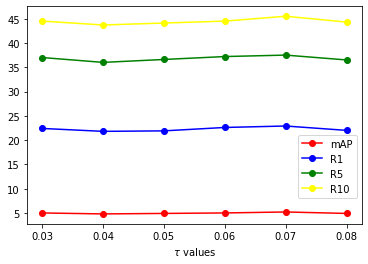}
\label{fig:tau_500authors}}
\hfil
\subfloat[Different $\lambda$ values on Market.]{\includegraphics[width=1.5in]{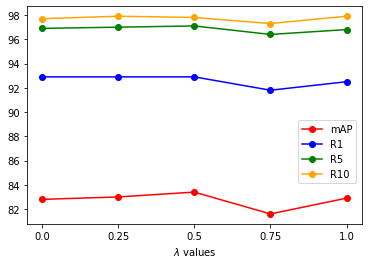}
\label{fig:lambda_market}}
\hfil
\subfloat[Different $\lambda$ values on Duke.]{\includegraphics[width=1.5in]{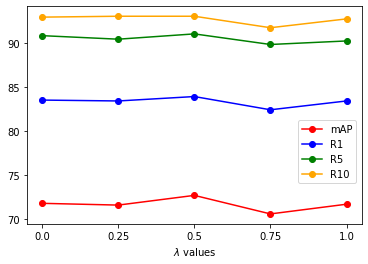}
\label{fig:lambda_duke}}
\hfil
\subfloat[Different $\lambda$ values on 50 authors.]{\includegraphics[width=1.5in]{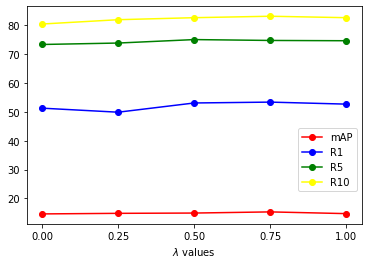}
\label{fig:lambda_50authors}}
\hfil
\subfloat[Different $\lambda$ values on 500 authors.]{\includegraphics[width=1.5in]{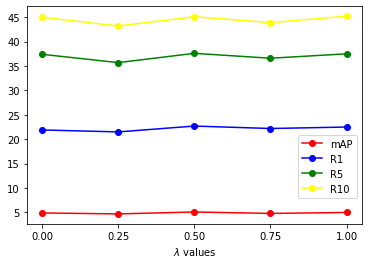}
\label{fig:lambda_500authors}}
\caption{Impact of different $\tau$ and $\lambda$ values on Market, Duke, 50 authors, and 500 authors datasets, considering mAP, R1, R5, and R10.}
\label{fig:lambda_tau_impact}
\end{figure}

The $\tau$ parameter, proposed in~\cite{hinton2015distilling}, is used to control the sharpness of the distribution related to the distance of a sample to each cluster proxy in the current iteration. The smaller the value, the greater the density towards the most confident value; while the larger the value, the closer the distribution is to the Uniform. Prior Unsupervised Person Re-Identification works usually tune it to control the gradients for a stable convergence. Results with varying $\tau$ are shown in Figures~\ref{fig:tau_market} and~\ref{fig:tau_duke} for ReID, and in Figures~\ref{fig:tau_50authors} and~\ref{fig:tau_500authors} for Authorship Attribution. For both datasets in ReID, the best value is $0.04$ as it provides the best mAP and R1, and top results for R5 and R10. For Authorship Attribution, the results have a marginal improvement for $\tau = 0.03$ in both 50 and 500 authors datasets, but it provides slightly lower performance for ReID. The greater values ($\tau = 0.07$) increase the loss and the gradient magnitude, which leads to suboptimal optimization for all datasets, mainly for ReID. Therefore, we set $\tau = 0.04$ for all experiments in both applications. 

Results with varying $\lambda$ (Equation \ref{eq:final_loss}) are shown in Figures~\ref{fig:lambda_market} and~\ref{fig:lambda_duke} for ReID, and in Figures~\ref{fig:lambda_50authors} and~\ref{fig:lambda_500authors} for Authorship Attribution. This hyper-parameter regulates the influence of the hard instance-based softmax-triplet loss  ($L_{hard}$) on the final loss. This loss brings a local view by considering hard-positive and -negative mining triplets at the batch level. When $\lambda = 0.0$ (i.e., $L_{hard}$ is not considered for optimization), we obtain one of the worst performances for all datasets in both modalities, showing that a more local view of the data is also important for model training. As we increase its value, the performance increases. But for higher values (from $0.75$ to $1.0$), we verify a performance drop as the local view starts dominating the global view, resulting in convergence issues. We keep $\lambda = 0.5$ in all experiments, as it yields the best or second-best mAP and R1 results for all datasets in both modalities.

\section{Conclusion}

In this work, we proposed a novel self-supervised learning pipeline for scenarios with high intra-class semantic disparity and inter-class similarity. General methods do not account for this problem as they are usually devised for less complex datasets, such as Imagenet.

Our pipeline starts from a common concept --- clustering steps to propose pseudo-labels for unlabeled samples and optimization steps to update backbones supervised by the pseudo-labels ---, but we incorporate novel techniques to address more critical tasks effectively. 

We propose the use of a neighborhood-based distance refinement followed by distance averaging to amalgamate complementary knowledge learned by different backbones. 
We showed that this is highly effective when compared to using distances obtained from each backbone directly. We provide a better distance measurement between samples, even without task-related initialization, due to the joint contribution of neighborhood-based distances and distance matrices ensemble.

Our second contribution is an ensemble-based clustering algorithm to provide pseudo-labels for optimization. 
The advantages are twofold: our solution creates dense but diverse clusters and does not need hyperparameter tuning. 

To show the generalizing ability of our pipeline, we applied it to two highly different Multimedia Forensics tasks: Person Re-Identification and Authorship Attribution from short text messages. To the best of our knowledge, this is the first self-supervised learning method that can be applied to different modalities with only minor adjustments.

For Person ReID, our method yields state-of-the-art performance in terms of mAP and Rank-1 in the most challenging datasets. For Authorship Attribution, we obtained competitive results when compared to a prominent method that was trained in a supervised manner. Therefore the ensemble-based clustering has a strong potential to find satisfactory clusters for model training on the fully-unsupervised scenario. Our self-supervised technique can considerably help the task of Textual Authorship Attribution in this forensic scenario, since it opens the possibility of using a massive amount of unlabeled data to foster the results.

We conclude that learning from complex fully-unlabeled data in different modalities is possible. Still, the model requires a robust distance measurement (brought by the ensemble of distance matrices) and a clustering strategy that tackles the unknown feature distribution from different datasets. When both strategies are put together, the method finds robust clusters for optimization.

One important aspect of the method that still needs optimization is memory usage. Currently, it requires quadratic memory $O(N^2)$, where $N$ is the total number of samples available for training, due to the pairwise distance matrices. Nonetheless, all prior art also faces the same issue.

Finally, our proposed method is applicable to a myriad of tasks, such as vehicle and place re-identification~\cite{khan2019survey, wang2020online}. In a future study, we aim to explore these new applications in an attempt to understand how the pipeline would adapt to new requirements. We would also like to consider different modalities simultaneously to propose possible connections among the main elements in a scene in a self-supervised way, contributing to scene or event understanding.


%



\section*{Acknowledgments}

We thank the financial support of the São Paulo Research Foundation (FAPESP), grants D\'ej\`aVu \#2017/12646-3, \#2019/15825-1, and \#2018/10204-6, and the support of Google through LARA (Latin America Research Awards).

\ifCLASSOPTIONcaptionsoff
  \newpage
\fi



\bibliographystyle{IEEEtran}
\bibliography{IEEEabrv, refs}
%


%

\begin{IEEEbiography}
[{\includegraphics[width=1in,height=3cm,clip,keepaspectratio]{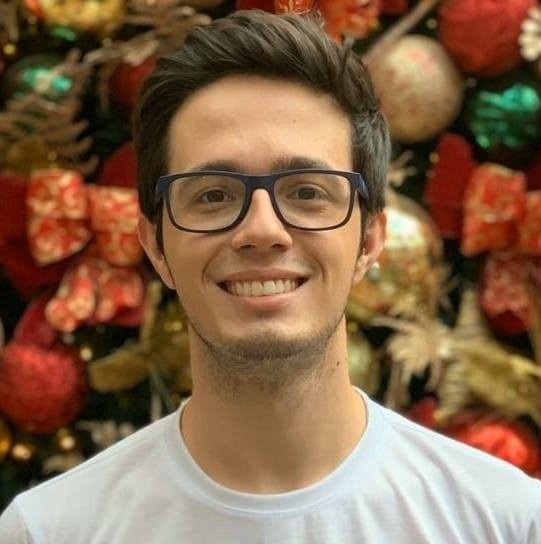}}]{Gabriel Bertocco} is currently pursuing his Ph.D. in Computer Science with a focus on digital forensics and machine learning at the Artificial Intelligence Lab. (\textbf{Recod.ai}) at the Institute of Computing, University of Campinas, Brazil, where he received a B.Sc. in Computing Engineering in 2019. His research interests include machine learning, computer vision, and digital forensics. Contact him at gabriel.bertocco@ic.unicamp.br.
\end{IEEEbiography}

\begin{IEEEbiography}[{\includegraphics[width=1in,height=1.25in,clip,keepaspectratio]{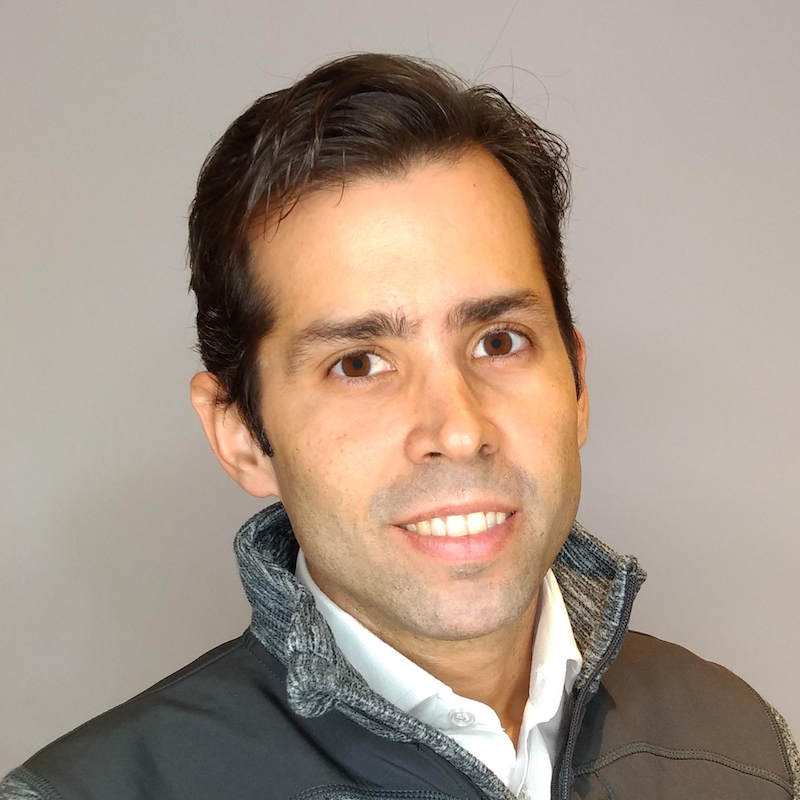}}]{Antonio Theophilo} is a researcher at the Center for Information Technology Renato Archer (CTI), Campinas, Brazil. He received his Ph.D. in Computer Science in 2022 at the Institute of Computing, University of Campinas, Brazil, and he is a researcher associated with the Artificial Intelligence Lab. (\textbf{Recod.ai}). His interests include Artificial Intelligence, Natural Language Processing, and Digital Forensics. He received the Google LARA (Latin America Research Awards) three times for his research over authorship attribution for small messages.
\end{IEEEbiography}

\begin{IEEEbiography}
[{\includegraphics[height=3cm,clip,keepaspectratio]{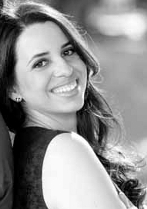}}]{Fernanda~Andal\'{o}} is a researcher associated with the Artificial Intelligence Lab. (\textbf{Recod.ai}) at the Institute of Computing, University of Campinas, Brazil. Andal\'{o} received a Ph.D. in Computer Science from the same university in 2012, during which she was a research fellow at Brown University. She worked for Samsung as a researcher and was a postdoctoral researcher in collaboration with Motorola, from 2014 to 2018. Currently, she works at The LEGO Group devising machine learning solutions for digital products. She was the 2016-2017 Chair of the IEEE Women in Engineering South Brazil Section, and is an elected member of the IEEE Information Forensics and Security Technical Committee. Her research interests include machine learning and computer vision.
\end{IEEEbiography}

\begin{IEEEbiography}
[{\includegraphics[height=3cm,clip,keepaspectratio]{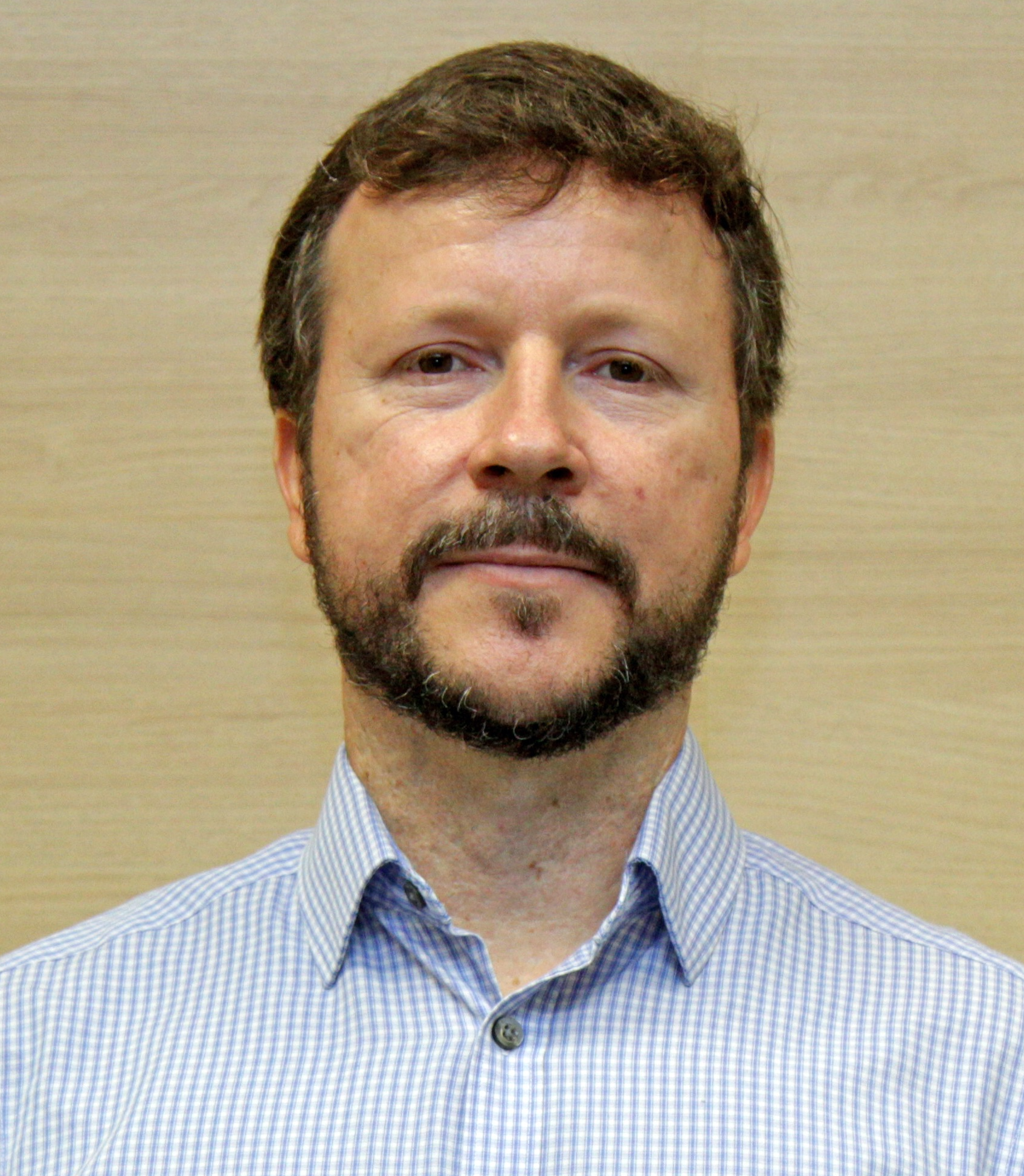}}]{Anderson Rocha} has been an associate professor at the Institute of Computing, the University of Campinas, Brazil, since 2009. Rocha received his Ph.D. in Computer Science from the University of Campinas. His research interests include Artificial Intelligence, Reasoning for complex data, and Digital Forensics. He is the Chair of the Artificial Intelligence Lab. (\textbf{Recod.ai}) at the Institute of Computing, University of Campinas. He was the Chair of the IEEE Information Forensics and Security Technical Committee for the 2019-2020 term. Finally, Prof. is an IEEE Senior Member, a Microsoft, Google and Tan Chi Tuan Faculty Fellow and is listed among the Top-1\% most influential scientists worldwide according to a study from Stanford Univ/Plos Biology. 
\end{IEEEbiography}




\end{document}


%
\title{Supplementary Material\\\vspace{0.7cm}\Large{\textbf{Leveraging Ensembles and Self-Supervised Learning for Fully-Unsupervised Person Re-Identification and Text Authorship Attribution}}}
%
%
%

\author{Gabriel~Bertocco,
        Antonio~Theophilo, 
        Fernanda~Andal\'{o},~\IEEEmembership{Member,~IEEE,}
        and~Anderson~Rocha,~\IEEEmembership{Senior~Member,~IEEE}}

\maketitle

%
\IEEEpeerreviewmaketitle

\section{Supplementary Material}
\label{sec:supmat}
%
%
%
%

\subsection{Comparison with Unsupervised Domain Adaptation}
Here we extend our comparison to prior art considering the works that employ a supervised pre-training on a source domain and then leverage their pipelines on the target domain. Our pipeline, as explained in the main article, does not require pre-training on task-related datasets and is applied directly over the target domain from weights initialized over ImageNet. Therefore, our model operates under a more challenging scenario with fewer constraints. The results are shown in Tables~\ref{tab:MarketAndDuke} and~\ref{tab:msmt17}.

\begin{table*}[!ht]
\caption{Results on Unsupervised Domain Adaptation. Our models \textbf{do not} require pre-training on the source domain. Here we present Market1501 to DukeMTMC-ReID and DukeMTMC-ReID to Market1501 adaptation scenarios. We highlight the three best models with \bv{blue}, \gv{green}, and \ov{orange}, respectively.}
\label{tab:MarketAndDuke}
\centering
\begin{tabular}{|p{2.7cm}| p{1.5cm}|p{0.8cm}|p{1.0cm}|p{1.0cm}|p{1.0cm}|p{0.8cm}|p{1.0cm}|p{1.0cm}|p{1.0cm}|}
\hline
\multicolumn{1}{|c|}{} &
\multicolumn{1}{|c|}{} &
\multicolumn{4}{|c|}{Duke $\rightarrow$ Market} & \multicolumn{4}{|c|}{Market $\rightarrow$ Duke} \\
\hline
Method & reference & mAP & R1 & R5 & R10
& mAP & R1 & R5 & R10 \\ \hline
PTGAN \cite{wei2018person} & CVPR'18 & - & 38.6 & - & 66.1 & - & 27.4 & - & 50.7 \\
PUL \cite{fan2018unsupervised} & TOMM'18 & 20.5 & 45.5 & 60.7 & 66.7 & 16.4 & 30.0 & 43.4 & 48.5\\
MMFA \cite{lin2018multi} & ArXiv'18 & 27.4 & 56.7 & 75.0 & 81.8 & 24.7 & 45.3 & 59.8 & 66.3\\
SPGAN \cite{deng2018image} & CVPR'18 & 22.8 & 51.5 & 70.1 & 76.8 & 22.3 & 41.1 & 56.6 & 63.0\\ 
TJ-AIDL \cite{wang2018transferable} & CVPR'18 & 26.5 & 58.2 & 74.8 & 81.1 & 23.0 & 44.3 & 59.6 & 65.0 \\
SPGAN+LMP \cite{deng2018image} & CVPR'18 & 26.7 & 57.7 & 75.8 & 82.4 & 26.2 & 46.4 & 62.3 & 68.0 \\
HHL \cite{zhong2018generalizing} & ECCV'18 & 31.4 & 62.2 & 78.8 & 84.0 & 27.2 & 46.9 & 61.0 & 66.7 \\
ATNet \cite{liu2019adaptive} & CVPR'19 & 25.6 & 55.7 & 73.2 & 79.4 & 24.9 & 45.1 & 59.5 & 64.2 \\
CamStyle \cite{zhong2018camstyle} & TIP'19 & 27.4 & 58.8 & 78.2 & 84.3 & 25.1 & 48.4 & 62.5 & 68.9 \\
MAR \cite{yu2019unsupervised} & CVPR'19 & 40.0 & 67.7 & 81.9 & - & 48.0 & 67.1 & 79.8 & -\\
PAUL \cite{yang2019patch} & CVPR'19 & 40.1 & 68.5 & 82.4 & 87.4 & 53.2 & 72.0 & 82.7 & 86.0 \\
ECN \cite{zhong2019invariance} & CVPR'19 & 43.0 & 75.1 & 87.6 & 91.6 & 40.4 & 63.3 & 75.8 & 80.4 \\
ISSDA-ReID \cite{tang2019unsupervised} & CVPR'19 & 63.1 & 81.3 & 92.4 & 95.2 & 54.1 & 72.8 & 82.9 & 85.9 \\
PDA-Net \cite{li2019cross} & ICCV'19 & 47.6 & 75.2 & 86.3 & 90.2 & 45.1 & 63.2 & 77.0 & 82.5 \\
CR-GAN \cite{chen2019instance} & ICCV'19 & 54.0 & 77.7 & 89.7 & 92.7 & 48.6 & 68.9 & 80.2 & 84.7 \\
PCB-PAST \cite{zhang2019self} & ICCV'19 & 54.6 & 78.4 & - & - & 54.3 & 72.4 & - & - \\
UCDA \cite{qi2019novel} & ICCV'19 & 30.9 & 60.4 & - & - & 31.0 & 47.7 & - & - \\
SSG \cite{fu2019self} & ICCV'19 & 58.3 & 80.0 & 90.0 & 92.4 & 53.4 & 73.0 & 80.6 & 83.2 \\
CASCL \cite{wu2019unsupervisedcascl} & ICCV'19 & 35.5 & 65.4 & 80.6 & 86.2 & 37.8 & 59.3 & 73.2 & 77.8 \\
UDAP \cite{song2020unsupervised} & PR'20 & 53.7 & 75.8 & 89.5 & 93.2 & 49.0 & 68.4 & 80.1 & 83.5 \\
MMCL \cite{wang2020unsupervised}& CVPR'20 & 60.4 & 84.4 & 92.8 & 95.0 & 51.4 & 72.4 & 82.9 & 85.0 \\
ACT \cite{yang2020asymmetric} & AAAI'20 & 60.6 & 80.5 & - & - & 54.5 & 72.4 & - & - \\
ECN-GPP \cite{zhong2020learning} & TPAMI'20 & 63.8 & 84.1 & 92.8 & 95.4 & 54.4 & 74.0 & 83.7 & 87.4 \\
SNR \cite{jin2020style} & CVPR'20 & 61.7 & 82.8 & - & - & 58.1 & 76.3 & - & - \\
AD-Cluster \cite{zhai2020ad} & CVPR'20 & 68.3 & 86.7 & 94.4 & 96.5 & 54.1 & 72.6 & 82.5 & 85.5 \\
MMT \cite{ge2020mutual} & ICLR'20 & 71.2 & 87.7 & 94.9 & \ov{96.9} & 65.1 & 78.0 & \ov{88.8} & 92.5 \\
DG-Net++ \cite{zou2020joint} & ECCV'20 & 61.7 & 82.1 & 90.2 & 92.7 & 63.8 & 78.9 & 87.8 & 90.4 \\
MEB-Net \cite{zhai2020multiple} & ECCV'20 & \ov{76.0} & \ov{89.9} & \ov{96.0} & \gv{97.5} & 66.1 & 79.6 & 88.3 & \ov{92.2} \\
ABMT \cite{chen2020enhancing} & WACV'20 & \gv{80.4} & \bv{93.0} & - & - & \ov{70.8} & \ov{83.3} & - & - \\
UST~\cite{9521886} & TIFS'21 & 78.4 & \gv{92.9} & \gv{96.9} & \bv{97.8} & \gv{72.6} & \bv{85.0} & \bv{92.1} & \bv{93.9} \\ \hline
\textbf{Ours} & \textbf{This work} & \bv{83.4} & \gv{92.9} & \bv{97.1} & \bv{97.8} & \bv{72.7} & \gv{83.9} & \gv{91.0} & \gv{93.0} \\ \hline 
\end{tabular}
\end{table*}

Despite not training on the source domain, we can see in Table~\ref{tab:MarketAndDuke} that our pipeline yields the best mAP on both scenarios and the first or at least the second best value for all other metrics. On the challenging MSMT17 (Table~\ref{tab:msmt17}), we present an even better performance by obtaining the best values for all metrics, surpassing the second-best method by a margin in all metrics. This shows the strong ability of our model to learn in a fully-unsupervised scenario without requiring any meta-information, source domain, or manual definition of hyper-parameter clustering.

\begin{table*}[ht]
\caption{Results on Unsupervised Domain Adaptation. We present Market1501 to MSMT17 and DukeMTMCRe-ID to MSMT17 adaptation scenarios. Our models \textbf{do not} require pre-training on the source domain so we replicate our results for both considered scenarios. We highlight the three best models with \bv{blue}, \gv{green}, and \ov{orange}, respectively.}
\label{tab:msmt17}
\centering
\begin{tabular}{|p{2.7cm}| p{2.0cm}|p{0.8cm}|p{1.0cm}|p{1.0cm}|p{1.0cm}|p{0.8cm}|p{1.0cm}|p{1.0cm}|p{1.0cm}|}
\hline
\multicolumn{1}{|c|}{} &
\multicolumn{1}{|c|}{} &
\multicolumn{4}{|c|}{Duke $\rightarrow$ MSMT17} & \multicolumn{4}{|c|}{Market $\rightarrow$ MSMT17} \\
\hline
Method & reference & mAP & R1 & R5 & R10
& mAP & R1 & R5 & R10 \\ \hline
PTGAN \cite{wei2018person} & CVPR 2018 & 3.3 & 11.8 & - & 27.4 & 2.9 & 10.2 & - & 24.4 \\
ECN \cite{zhong2019invariance} & CVPR 2019 & 10.2 & 30.2 & 41.5 & 46.8 & 8.5 & 25.3 & 36.3 & 42.1 \\
SSG \cite{fu2019self} & ICCV 2019 & 13.3 & 32.2 & - & 51.2 & 13.2 & 31.6 & - & 49.6 \\
ECN-GPP \cite{zhong2020learning} & TPAMI 2020 & 16.0 & 42.5 & 55.9 & 61.5 & 15.2 & 40.4 & 53.1 & 58.7 \\
MMCL \cite{wang2020unsupervised} & CVPR 2020 & 16.2 & 43.6 & 54.3 & 58.9 & 15.1 & 40.8 & 51.8 & 56.7 \\
MMT \cite{ge2020mutual} & ICLR 2020 & 23.3 & 50.1 & 63.9 & 69.8 & 22.9 & 49.2 & 63.1 & 68.8 \\
DG-Net++  \cite{zou2020joint} & ECCV 2020 & 22.1 & 48.8 & \ov{60.9} & \ov{65.9} & 22.1 & 48.4 & 60.9 & 66.1 \\
ABMT \cite{chen2020enhancing} & WACV 2020 & \ov{33.0} & \ov{61.8} & - & - & 27.8 & 55.5 & - & - \\ 
SpCL \cite{ge2020self} & NeurIPS 2020 & - & - & - & - & \ov{31.0} & \ov{58.1} & \ov{69.6} & \ov{74.1} \\
UST~\cite{9521886} & TIFS'21  & \gv{34.5} & \gv{63.9} & \gv{75.3} & \gv{79.6} & \gv{33.2} & \gv{62.3} & \gv{74.1} & \gv{78.5} \\ \hline
\textbf{Ours} & \textbf{This work} & \bv{42.6} & \bv{68.2} & \bv{77.9} & \bv{81.4} & \bv{42.6} & \bv{68.2} & \bv{77.9} & \bv{81.4}\\ \hline
\end{tabular}
\end{table*}

\subsection{Further Qualitative Analysis}

In the main article, we show the qualitative analysis considering one success and one failure cases for Market and Duke datasets. We extend this analysis here and show the activated regions in the gallery images given a query from each camera in each dataset, considering the ResNet50 backbone. Blue frames indicate the query image, green indicates the true positive samples, and red the false positive samples. We adapted the method from~\cite{stylianou2019visualizing} to visualize the results.

\begin{figure*}[ht]
\centering
\subfloat[Camera 1]{\includegraphics[width=3.5in]{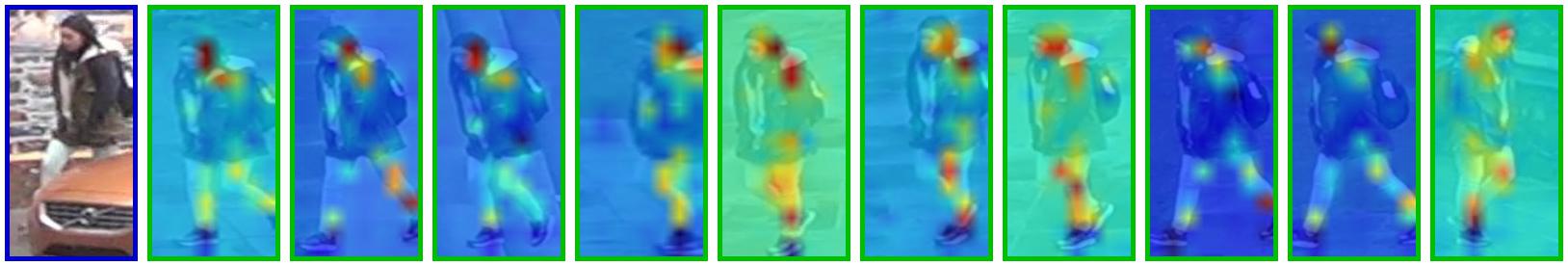}
\label{fig:M2D_successful_cam01}}
\hfil
\subfloat[Camera 2]{\includegraphics[width=3.5in]{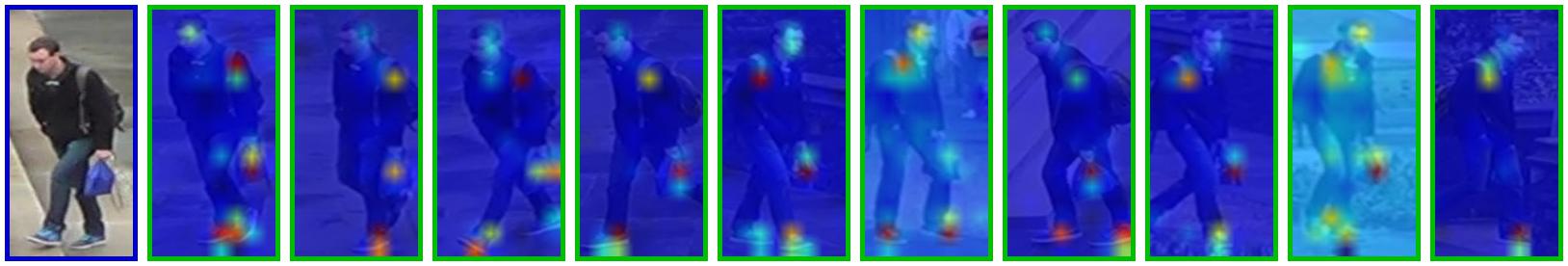} 
\label{fig:M2D_successful_cam02}}
\hfil
\subfloat[Camera 3]{\includegraphics[width=3.5in]{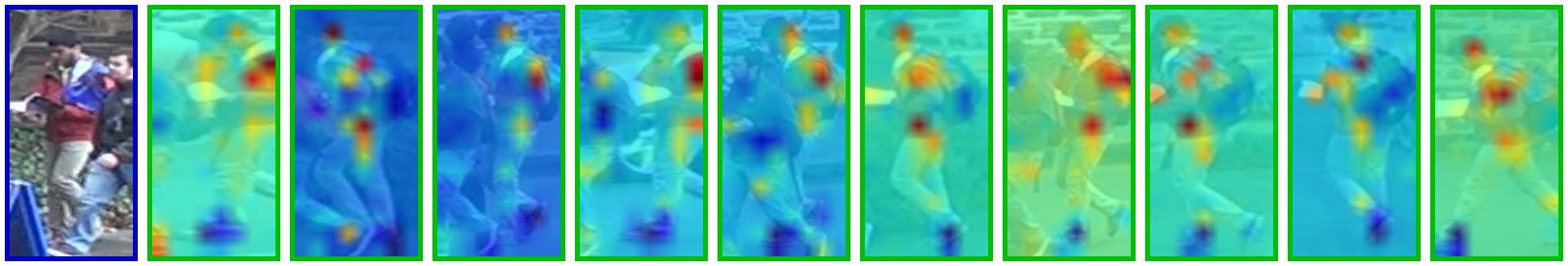} 
\label{fig:M2D_successful_cam03}}
\hfil
\subfloat[Camera 4]{\includegraphics[width=3.5in]{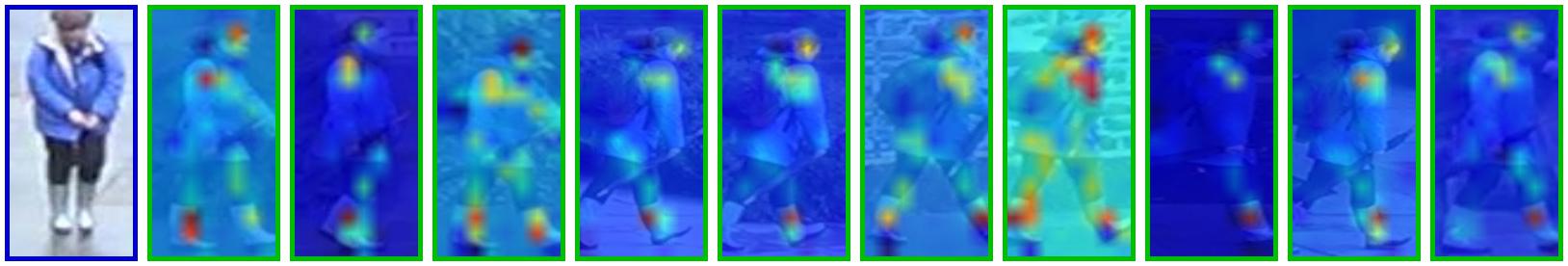} 
\label{fig:M2D_successful_cam04}}
\hfil
\subfloat[Camera 5]{\includegraphics[width=3.5in]{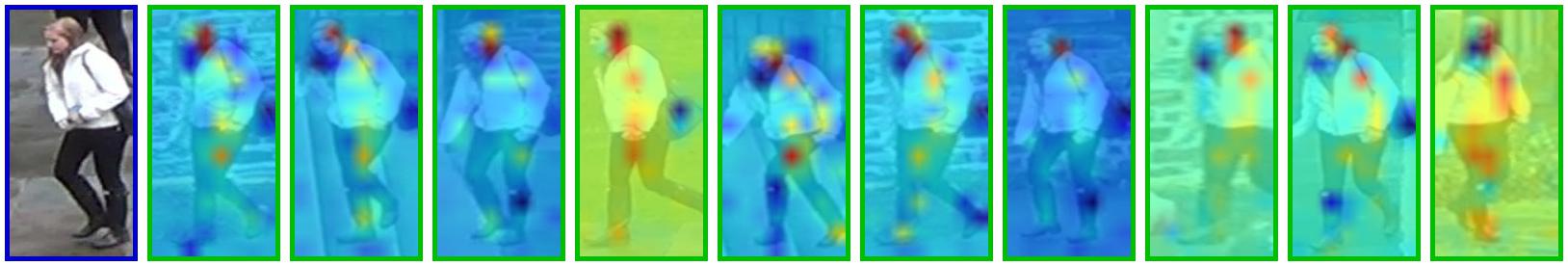} 
\label{fig:M2D_successful_cam05}}
\hfil
\subfloat[Camera 6]{\includegraphics[width=3.5in]{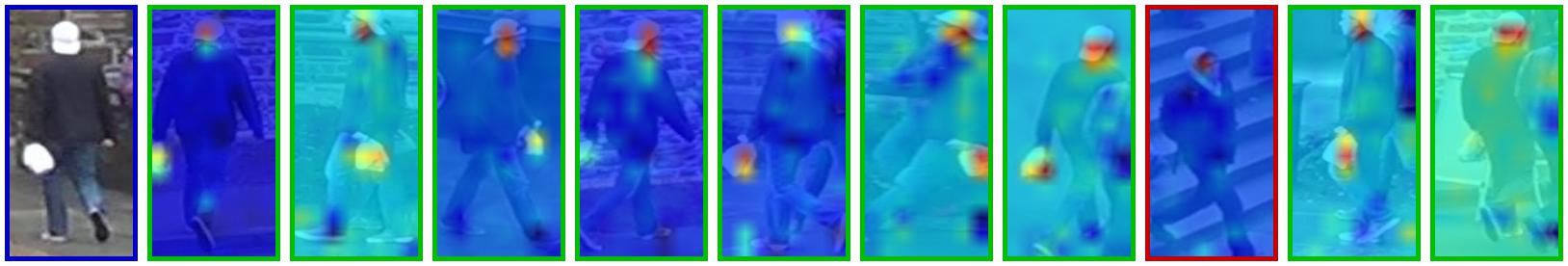} 
\label{fig:M2D_successful_cam06}}
\hfil
\subfloat[Camera 7]{\includegraphics[width=3.5in]{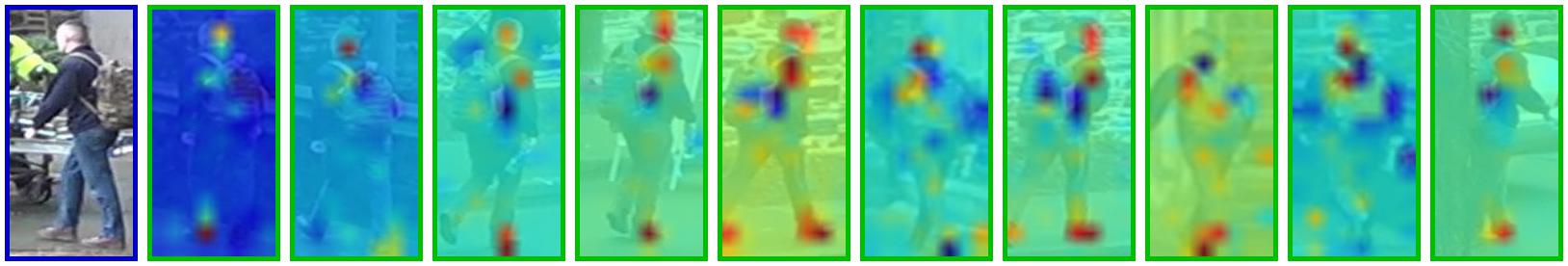} 
\label{fig:M2D_successful_cam07}}
\hfil
\subfloat[Camera 8]{\includegraphics[width=3.5in]{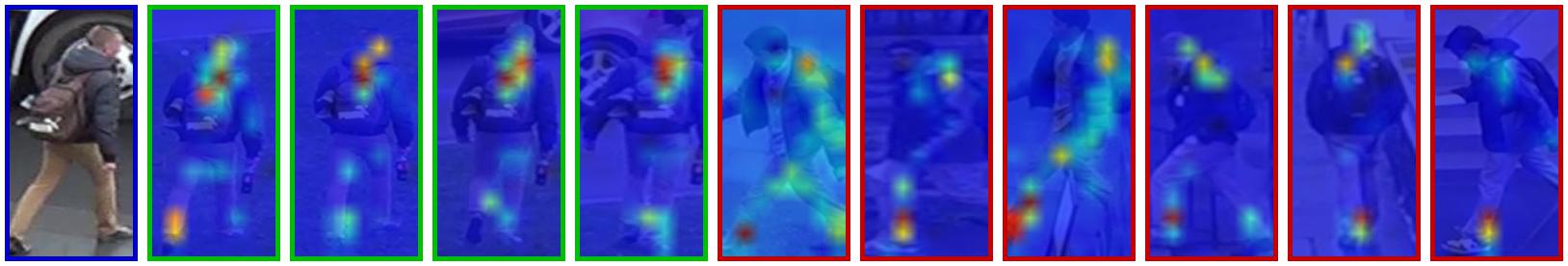} 
\label{fig:M2D_successful_cam08}}
\caption{Success cases considering one query from each camera on Duke. These results are obtained with the ResNet50 backbone after training on Duke.}
\label{fig:M2D_successful_cases}
\end{figure*}

\begin{figure*}[ht]
\centering
\subfloat[Camera 1]{\includegraphics[width=3.5in]{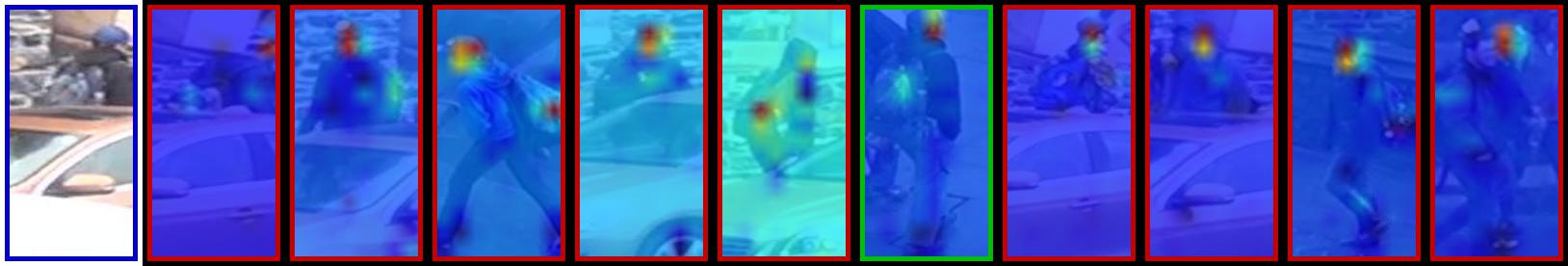}
\label{fig:M2D_failure_cam01}}
\hfil
\subfloat[Camera 2]{\includegraphics[width=3.5in]{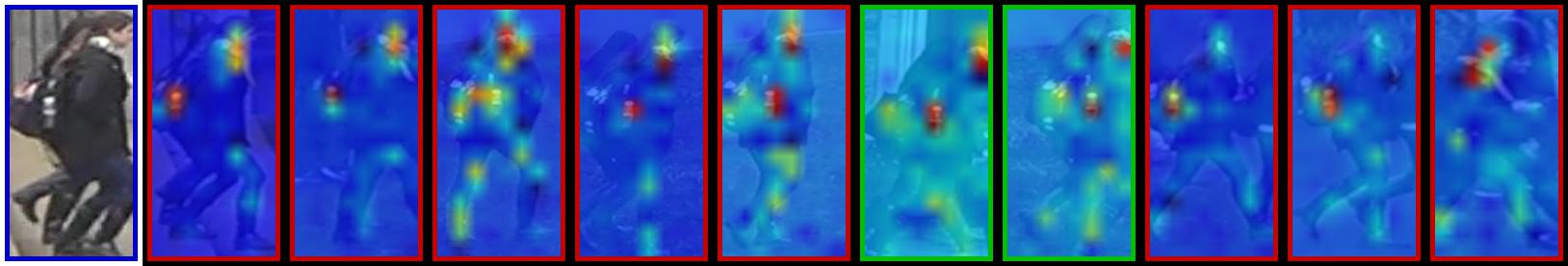} 
\label{fig:M2D_failure_cam02}}
\hfil
\subfloat[Camera 3]{\includegraphics[width=3.5in]{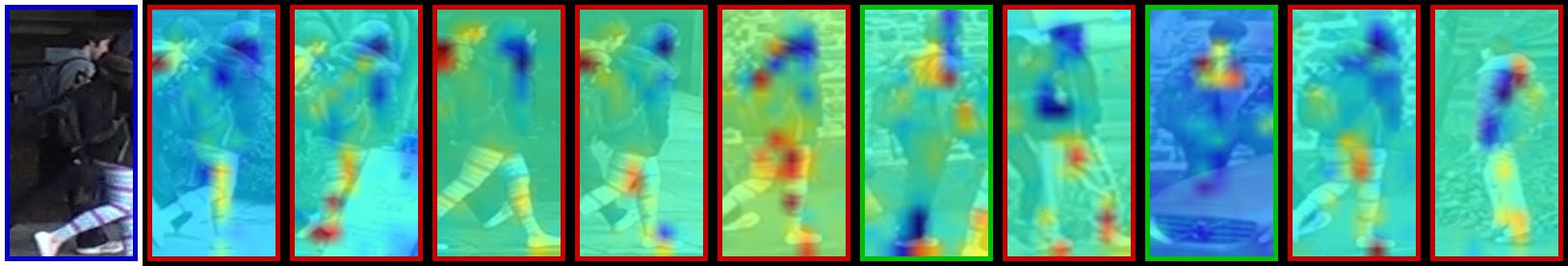} 
\label{fig:M2D_failure_cam03}}
\hfil
\subfloat[Camera 4]{\includegraphics[width=3.5in]{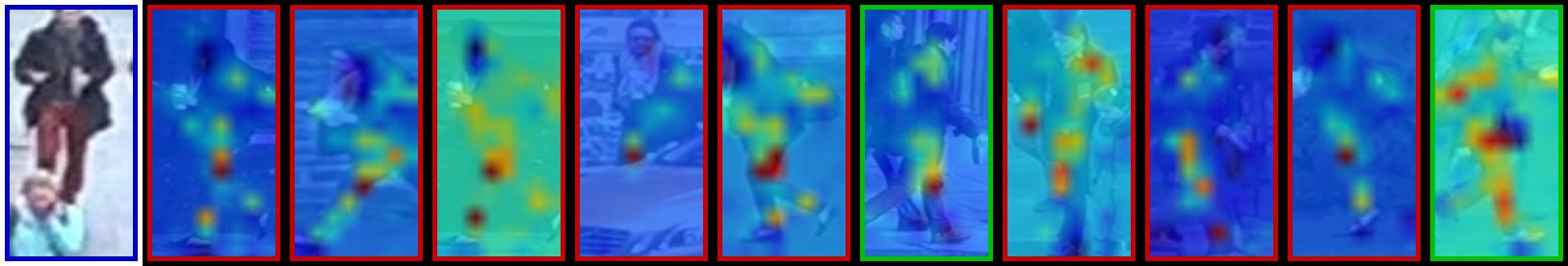} 
\label{fig:M2D_failure_cam04}}
\hfil
\subfloat[Camera 5]{\includegraphics[width=3.5in]{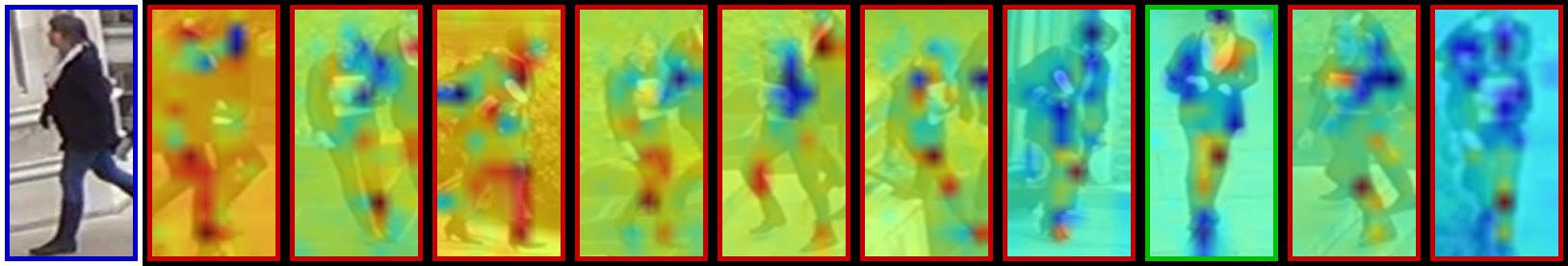} 
\label{fig:M2D_failure_cam05}}
\hfil
\subfloat[Camera 6]{\includegraphics[width=3.5in]{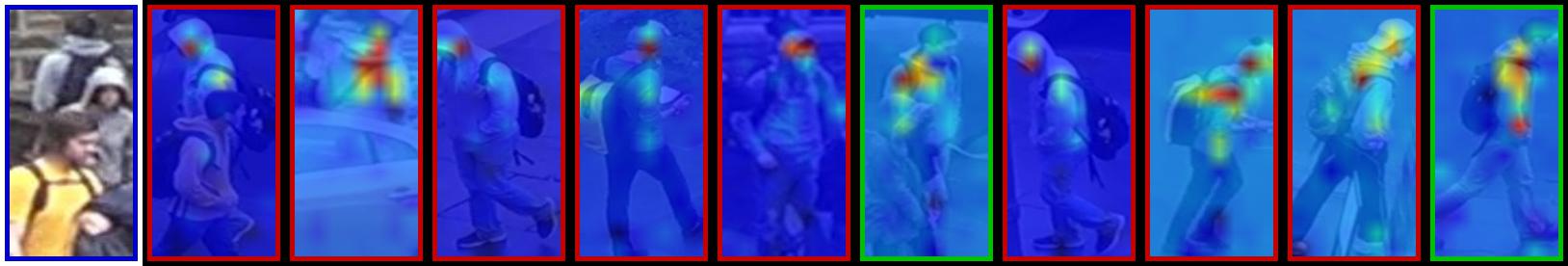} 
\label{fig:M2D_failure_cam06}}
\hfil
\subfloat[Camera 7]{\includegraphics[width=3.5in]{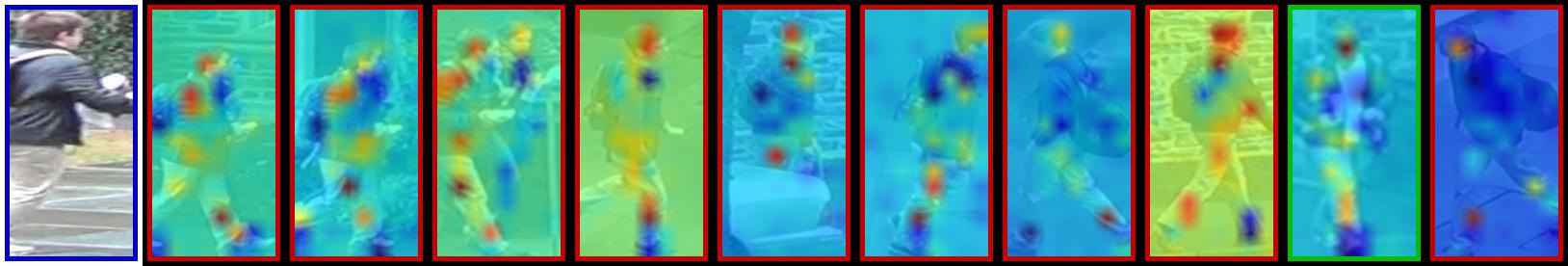} 
\label{fig:M2D_failure_cam07}}
\hfil
\subfloat[Camera 8]{\includegraphics[width=3.5in]{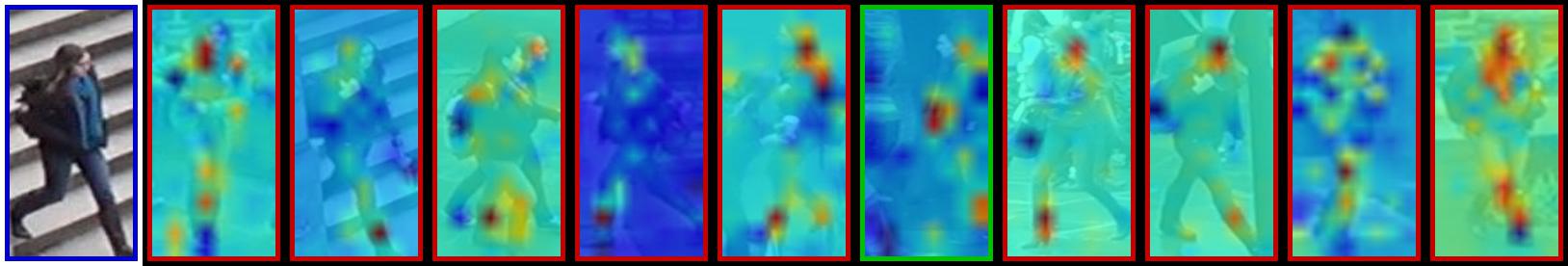} 
\label{fig:M2D_failure_cam08}}
\caption{Failure cases considering one query from each camera on Duke. These results are obtained with the ResNet50 backbone after training on Duke.}
\label{fig:M2D_failure_cases}
\end{figure*}

Figure~\ref{fig:M2D_successful_cases} shows one success case for each one of the eight cameras present on the Duke dataset. We see that the model is able to mine fine-grained details on all images, regardless of the camera, and retrieve true positive samples in the top 10 images. The only exception is for camera 8 (Figure~\ref{fig:M2D_successful_cam08}), where the last six images are false positive samples. However, for all samples, the model still focuses on fine-grained details, and the errors occur due to similar clothing of the identities. The same happens in the examples in Figure~\ref{fig:M2D_failure_cases}. Despite being error cases, the model can focus on fine-grained details. The errors occur when there are two or more identities in the query image (Figure~\ref{fig:M2D_failure_cam06}), the background color is similar to an object held by the identity (in Figure~\ref{fig:M2D_failure_cam03} the backpack is the same color of the background), or due to very similar clothing of different identities. Since Duke was recorded at a university campus during winter, it is natural to find individuals wearing similar (and dark) clothing.

\begin{figure*}[ht]
\centering
\subfloat[Camera 1]{\includegraphics[width=3.5in]{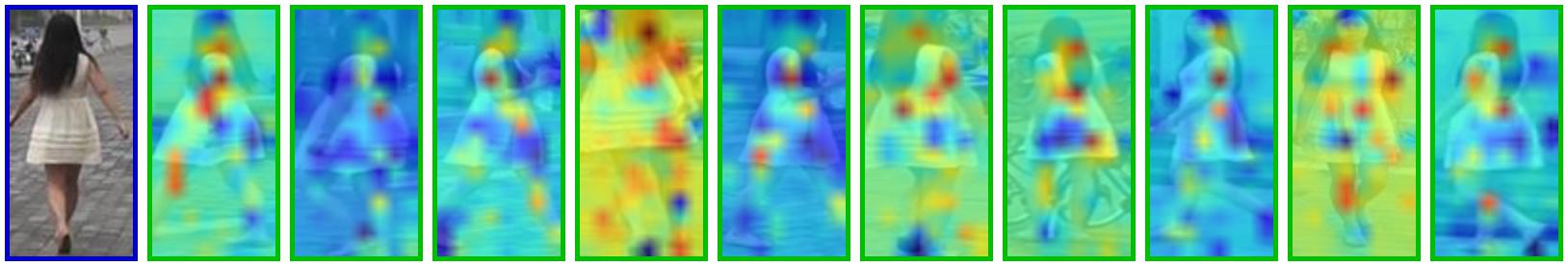}
\label{fig:D2M_successful_cam01}}
\hfil
\subfloat[Camera 2]{\includegraphics[width=3.5in]{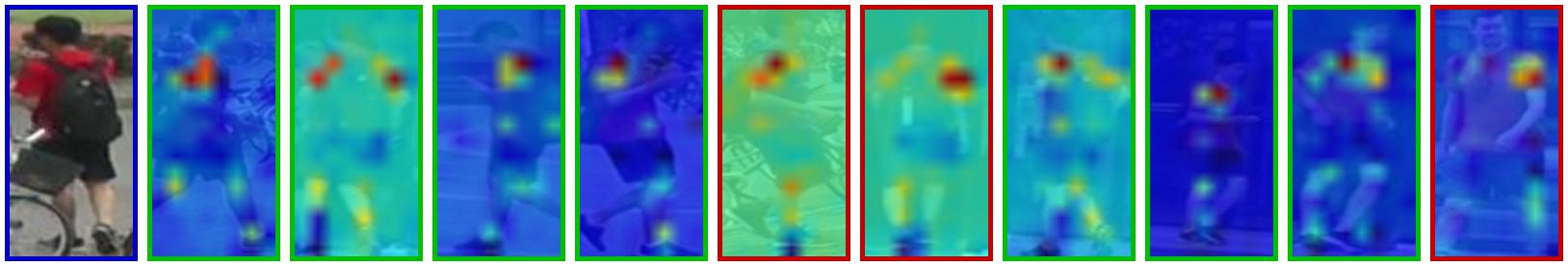} 
\label{fig:D2M_successful_cam02}}
\hfil
\subfloat[Camera 3]{\includegraphics[width=3.5in]{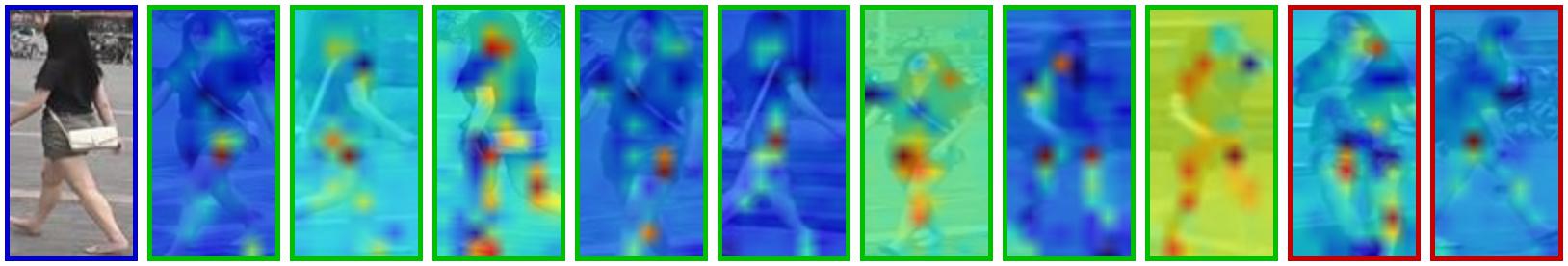} 
\label{fig:D2M_successful_cam03}}
\hfil
\subfloat[Camera 4]{\includegraphics[width=3.5in]{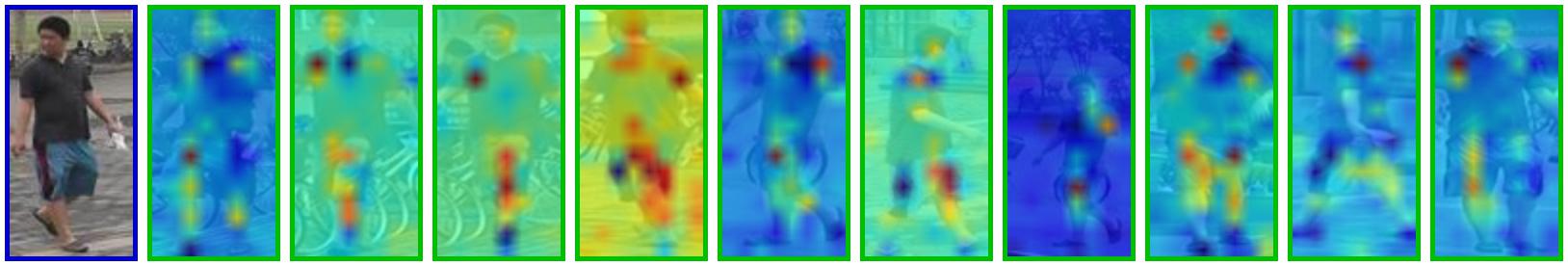} 
\label{fig:D2M_successful_cam04}}
\hfil
\subfloat[Camera 5]{\includegraphics[width=3.5in]{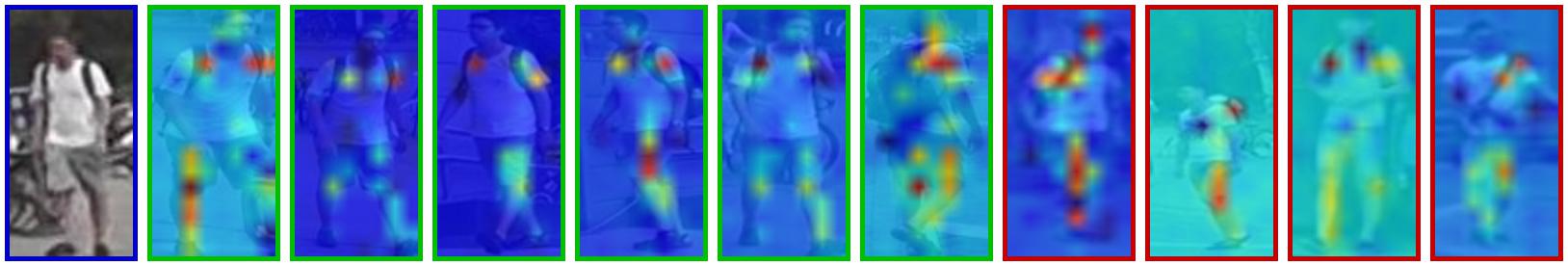} 
\label{fig:D2M_successful_cam05}}
\hfil
\subfloat[Camera 6]{\includegraphics[width=3.5in]{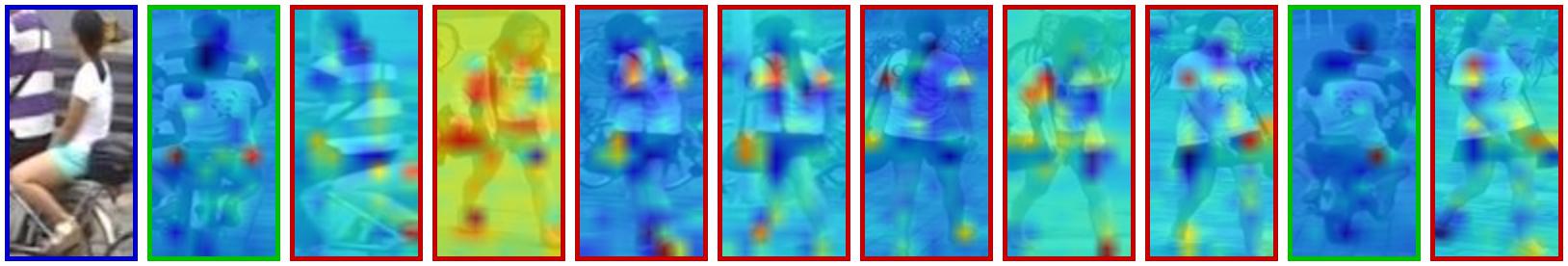} 
\label{fig:D2M_successful_cam06}}
\caption{Success cases considering one query from each camera on Market. These results are obtained with the ResNet50 backbone after training on Market.}
\label{fig:D2M_successful_cases}
\end{figure*}

\begin{figure*}[ht]
\centering
\subfloat[Camera 1]{\includegraphics[width=3.5in]{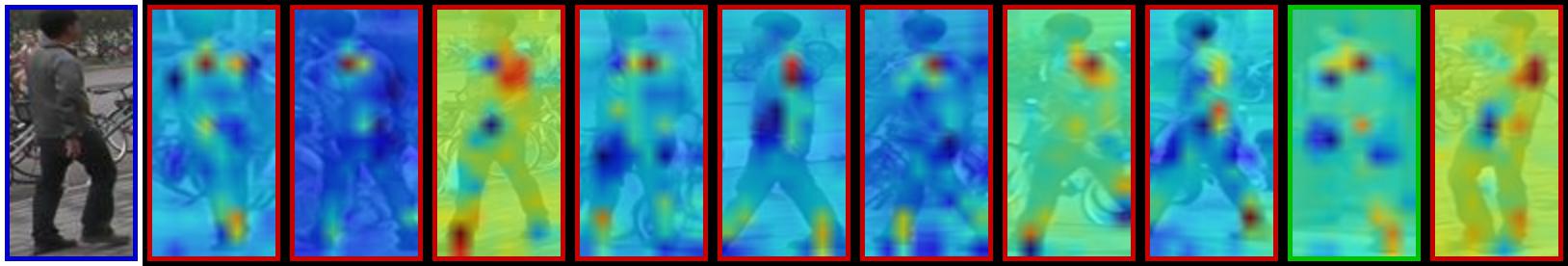}
\label{fig:D2M_failure_cam01}}
\hfil
\subfloat[Camera 2]{\includegraphics[width=3.5in]{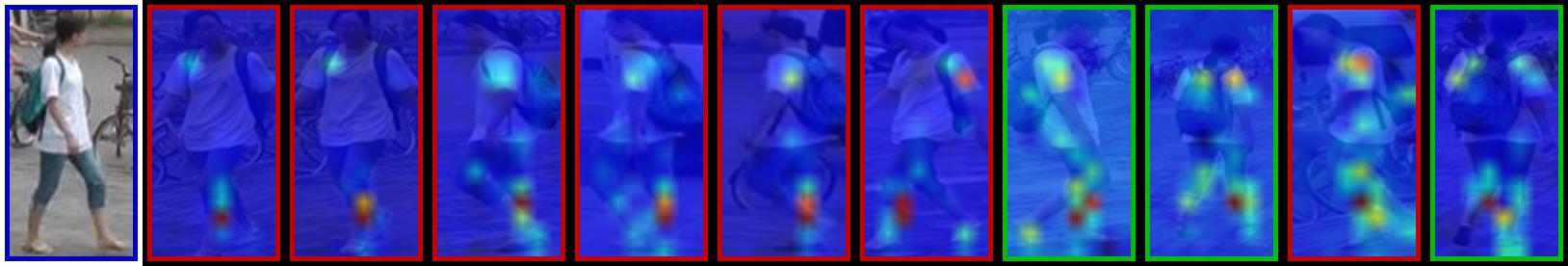} 
\label{fig:D2M_failure_cam02}}
\hfil
\subfloat[Camera 3]{\includegraphics[width=3.5in]{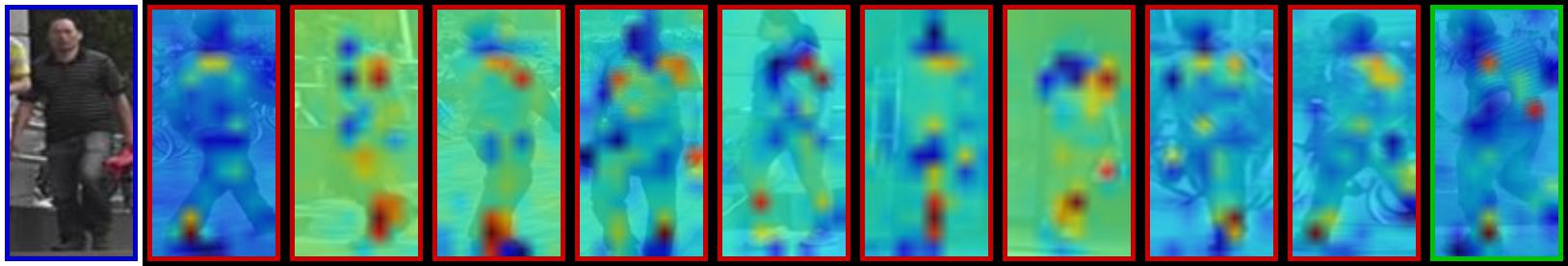} 
\label{fig:D2M_failure_cam03}}
\hfil
\subfloat[Camera 4]{\includegraphics[width=3.5in]{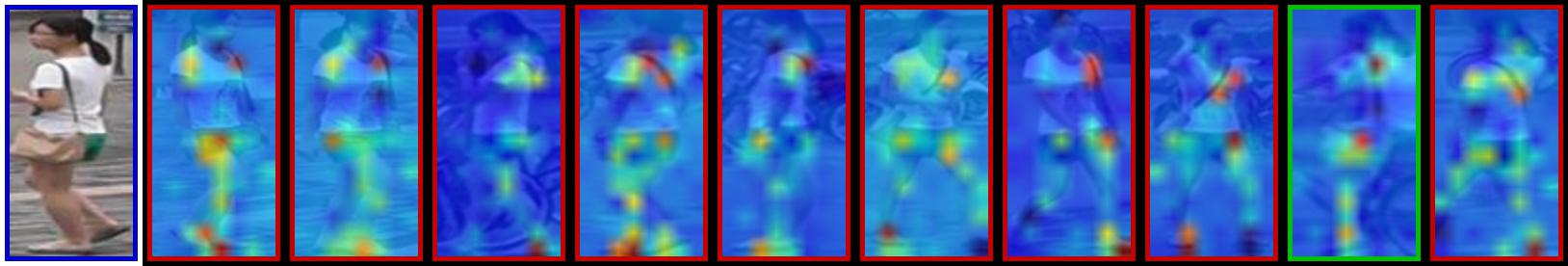} 
\label{fig:D2M_failure_cam04}}
\hfil
\subfloat[Camera 5]{\includegraphics[width=3.5in]{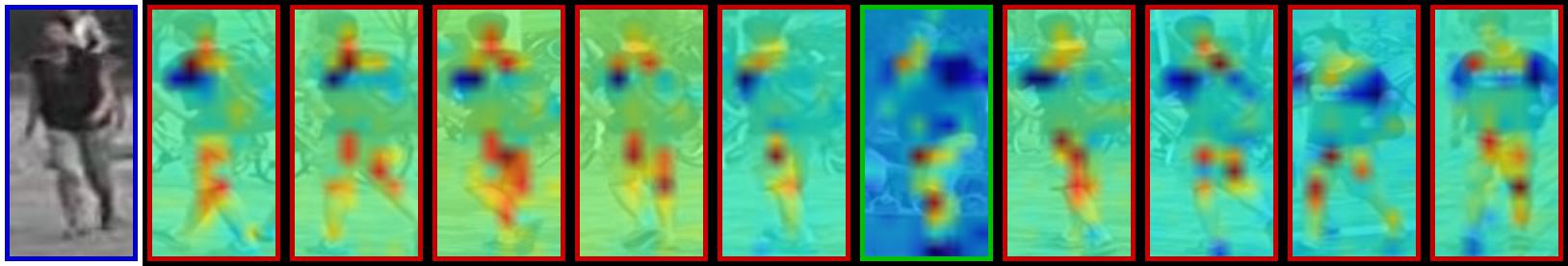} 
\label{fig:D2M_failure_cam05}}
\hfil
\subfloat[Camera 6]{\includegraphics[width=3.5in]{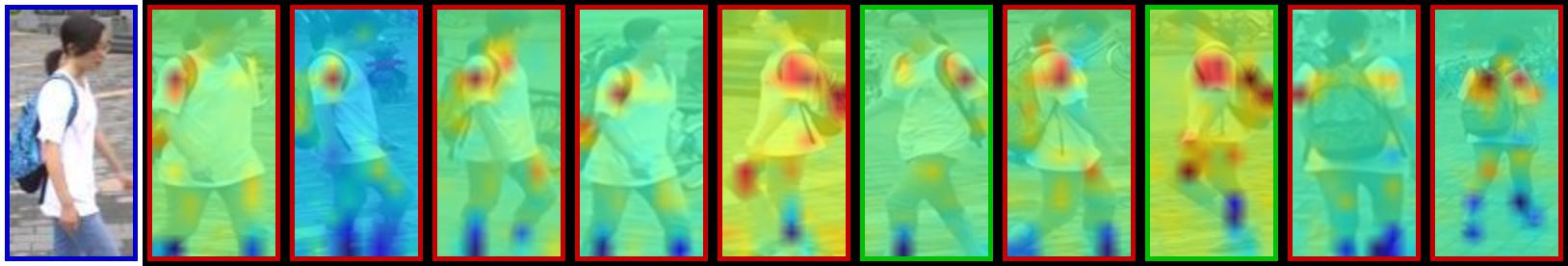} 
\label{fig:D2M_failure_cam06}}
\caption{Failure cases considering one query from each camera on Market. These results are obtained with the ResNet50 backbone after training on Market.}
\label{fig:D2M_failure_cases}
\end{figure*}

Figure~\ref{fig:D2M_successful_cases} shows one success case for each camera on the Market dataset. Following the same previous conclusions, the model is able to identify fine-grained features in the image to correctly retrieve the true positive samples among the top 10 images. The failure cases (Figure~\ref{fig:D2M_failure_cases}) are mainly due to the identities being similar (similar T-shirts, hair, and skin tone), which is the case in Figures~\ref{fig:D2M_failure_cam02}, \ref{fig:D2M_failure_cam04}, and~\ref{fig:D2M_failure_cam06}.

To summarize, the model is able to identify fine-grained details to match the same identity, allowing us to outperform prior art.

\subsection{Further Analysis}

Our applications are Unsupervised Person Re-Identification and Unsupervised Text Authorship Attribution, which are related to event analysis in a forensic context. Thus, it is important to understand how the performance of our approach is impacted when we consider common forensics features used for other tasks, such as image manipulation detection~\cite{barni2020improving}, or pre-processed inputs used in the GAN-generated image detection task~\cite{barni2020cnn}. 

To test this, we apply the method proposed in~\cite{barni2020improving} for manipulation detection to the Unsupervised Person Re-Identification problem. It is an ensemble-based algorithm with forensics features. More specifically, the authors proposed the use of a 1.5C classifier, which is composed of one two-class SVM and three one-class SVMs, achieving top-tier performance on the task. 
We employ the same features based on SPAM features~\cite{5437325}, and the CRSPAM1372~\cite{barni2018detection} features which have also shown to achieve top-tier performance on the manipulation detection task.

However, since Unsupervised Person Re-Identification differs in nature from manipulation detection, SPAM and CRSPAM1372 features induce a huge performance drop when employed in our setup. The main reasons are:
\begin{enumerate}
    \item These features are hand-crafted and were not designed for classification problems dependent on semantic context. They are not able to overcome the non-linearity present in Unsupervised Person Re-Identification problems caused by the difference in illumination, resolution, pose, background, and occlusion. Moreover, those features are designed to highlight unusual or noise patterns in the relationship between neighboring pixels, which describe the signal signature on images that goes through some manipulation. For this reason, they do not describe the semantics of the image (e.g., shoes, arms, body, legs, clothes), which are fundamental to performing the cross-camera retrieval task in Person Re-Identification. Our model, on the other hand, does not focus on neighboring pixels inconsistencies but on learning high-level semantic features. Consequently, it does not learn features sensitive to neighboring pixels differences caused by noise or camera signature but, on the contrary, it is invariant to them since the main goal of the Unsupervised Person Re-Identification is to match the same person across cameras. 

    \item The output of the SVM is a classification score or the distance to the decision boundary, since it aims to project the data in a linearly separable feature space and perform classification in a single step. On the other hand, the DCNN has multiple feature space transformations by forwarding the data from one layer to another, projecting the intermediate features to higher semantic levels, easing the verification task.

    \item Our main goal is to progressively find pseudo-classes in fully-unlabeled data to learn high-level and general features. In this context, the one-class SVM is not applicable since its main goal is to create a decision boundary (hyper-ball) around the samples of interest, and not to find pseudo-classes and progressively learn from them.
\end{enumerate}

We also consider pre-processing techniques employed in the GAN-generated image detection task. We take the method proposed in~\cite{barni2020cnn}, whose main goal is distinguishing between GAN-generated and real images. To generate a digital image, most CCD and CMOS camera sensors capture the light, and the digital system performs processing and interpolation to create pixel values for each of the three channels (R, G, and B). This process creates a specific and unique signature in the image related to the acquisition sensor. When synthesizing images with GANs, the generator aims to create an image as close as possible to a real one by imitating the neighboring interpolation (e.g., mosaic and demosaic process) and the camera signature (e.g., Photo-Response Non-Uniformity -- PRNU~\cite{9599116}) present in the real image. However, this process is not perfect and leaves traces that are used to perform GAN-generated image detection.

In~\cite{barni2020cnn}, the authors argue that it is harder for GANs to reconstruct consistent relationships among the color bands, and they propose an approach to detect these inconsistencies. Their method outputs a six-channel matrix which is the input image for our model, where the first three channels correspond to the intra-color-band pixel co-occurrence (for channels R, G, and B), and the last three to the cross-color-band pixel co-occurrences. For a fair comparison, we keep our whole pipeline and change only the input of our backbones to match the ones from~\cite{barni2020cnn}. Their method induces a huge performance drop compared to ours with standard RGB images. The mAP for their model in the three datasets faced a performance lower than $10.0\%$, while our lowest mAP is $42.6\%$ in MSMT17. All of the ranking metrics with their method also do not surpass $20.0\%$, while our lowest ranking metric is $68.2\%$ in MSMT17. The main reason is the difference in the nature of the tasks addressed by the methods. Their method requires a signal analysis across color bands seeking the reconstruction of inconsistencies and aiming to detect the neighboring inconsistencies and noise introduced by GAN-based models. However, this process might destroy the semantics present in the images, which is fundamental in Unsupervised Person Re-Identification.
Our models learn high-level semantic and camera-invariant features to match the same person seen from different camera views and distinguish from other people by looking at discriminant parts, such as clothes, shoes, bags, face, etc. (Figures~\ref{fig:M2D_successful_cases} and~\ref{fig:D2M_successful_cases}). In summary, the method proposed in~\cite{barni2020cnn} looks for fine-grained details at the noise level, while ours looks for fine-grained details at the semantic level. 


%



\ifCLASSOPTIONcaptionsoff
  \newpage
\fi



\bibliographystyle{IEEEtran}
\bibliography{IEEEabrv, refs}
%


%



